\documentclass[sigconf]{acmart}

\usepackage{subfigure,multirow}
\usepackage{graphicx}
\usepackage{amsmath}
\usepackage{mathrsfs} 
\usepackage{algorithm}
\usepackage[noend]{algpseudocode}
\usepackage{booktabs}  

\definecolor{commentcolor}{RGB}{0, 128, 0} 

\copyrightyear{2025}
\acmYear{2025}
\setcopyright{cc}
\setcctype{by-nc}
\acmConference[MobiSys '25]{The 23rd Annual International Conference on
	Mobile Systems, Applications and Services}{June 23--27, 2025}{Anaheim, CA,
	USA}
\acmBooktitle{The 23rd Annual International Conference on Mobile Systems,
	Applications and Services (MobiSys '25), June 23--27, 2025, Anaheim, CA, USA}
\acmDOI{10.1145/3711875.3729132}
\acmISBN{979-8-4007-1453-5/2025/06}



\begin{document}

\title{Never Start from Scratch: Expediting On-Device LLM Personalization via Explainable Model Selection}

\author{Haoming Wang, Boyuan Yang, Xiangyu Yin, and Wei Gao}
\affiliation{
	\institution{University of Pittsburgh}
	\country{USA}
}
\email{{hw.wang, by.yang, eric.yin, weigao}@pitt.edu}

\begin{abstract}
Personalization of Large Language Models (LLMs) is important in practical applications to accommodate the individual needs of different mobile users. Due to data privacy concerns, LLM personalization often needs to be locally done at the user's mobile device, but such on-device personalization is constrained by both the limitation of on-device compute power and insufficiency of user's personal data. In this paper, we address these constraints by fine-tuning an already personalized LLM with user's personal data, and present XPerT, a new technique that ensure proper selection of such already personalized LLMs based on explainability about how they were being fine-tuned. We implemented and evaluated XPerT on various smartphone models with mainstream LLMs, and experiment results show that XPerT reduces the computation costs of on-device LLM personalization by 83\%, and improves its data efficiency by 51\%.
\end{abstract}

\maketitle

\vspace{-0.05in}
\section{Introduction}

Large Language Models (LLMs) have demonstrated unprecedented capabilities in natural language understanding and generation, enabling intelligent applications in diverse domains. To better meet the individualized needs of different mobile users, LLM personalization is used in practice \cite{brown2020language,chung2024scaling}, by fine-tuning the LLM with users' personal data and feedback \cite{tseng2024two,zhang2024personalization,huang2024towards,huang2025mpnp} generated from their mobile devices. For example, a personalized LLM-driven agent on smartphones provides better task automations and content suggestions based on user's preferences and contexts \cite{park2023generative,xie2024large,huang2024perceptual}, and better meets the user's need of specific language styles \cite{li2024personal,yin2023ptease}.


To personalize a LLM, a naive approach is to upload personal data to the cloud server where the LLM is fine-tuned \cite{cai2024edge,ding2024enhancing}. However, 
exposing the user's personal data could seriously impair the data privacy \cite{yao2024survey}. Instead, 
a better option is to fine-tune the LLM at the user's local mobile device \cite{huang2023elastictrainer}, but such on-device LLM personalization is constrained by both the limitation of on-device compute power and insufficiency of personal data \cite{huang2022real}. First, 
fine-tuning LLMs, even those small ones (e.g., Llama3.2-1B \cite{dubey2024llama} and Qwen2-0.5B \cite{yang2024qwen2}), is expensive on mobile devices and incurs long latency. For example, one training step of fine-tuning the Qwen2-0.5B model on a flagship smartphone, such as Google Pixel 9 Pro, need at least hundreds of milliseconds, even with parameter-efficient fine-tuning methods such as LoRA \cite{hu2021lora}. Second, effective LLM fine-tuning requires large amounts of training data. Usually, at least thousands of data samples are required to avoid bias and overfitting \cite{zhang2024scaling}, but accumulating such amount of on-device data takes very long time\footnote{For example, a smartphone user daily sends and receives only about 50-100 text messages on average \cite{smith2019americans}.}. Continual learning can allow LLM fine-tuning in the meantime \cite{razdaibiedina2023progressive,wang2023trace}, but is also too expensive for mobile devices \cite{shi2024continual}.


\begin{figure}
	\centering
	\includegraphics[width=2.5in]{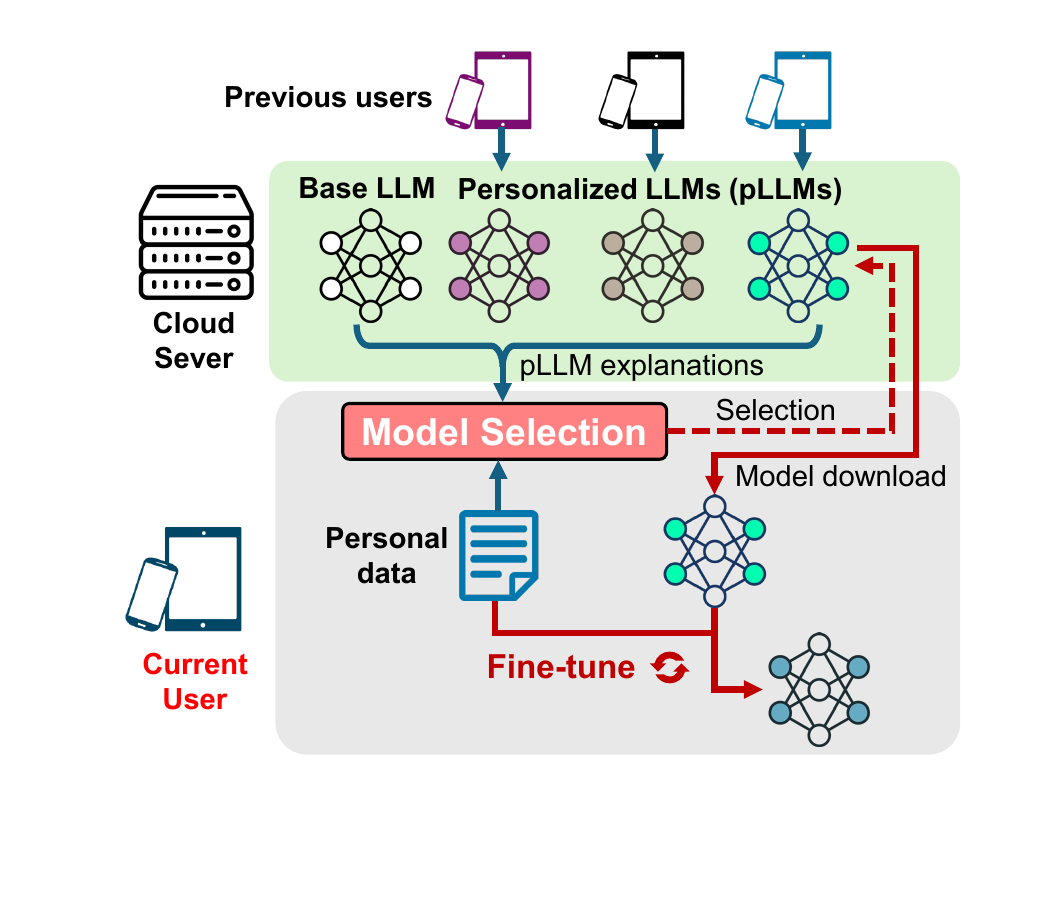}
	\vspace{-0.15in}
	\caption{On-device LLM personalization via explainable model selection}
	\label{fig:framework_overview}
	\vspace{-0.15in}
\end{figure}

To address these challenges and ensure prompt on-device LLM personalization, in this paper we advocate a fundamentally different approach: as shown in Figure \ref{fig:framework_overview}, every time after a LLM has been personalized at a user's device, the user uploads its personalized LLM, namely ``\emph{pLLM}'', to be cached at the cloud server. Afterwards, when a current user performs on-device LLM personalization, instead of fine-tuning a pre-trained base LLM from scratch, it checks the cloud server for a cached pLLM that best approximates to the current user's personalization need. If such a cached pLLM is found, the current user downloads and fine-tunes this cached pLLM using its personal data. The number of training epochs and the amount of training data required for fine-tuning the cached pLLM can then be largely reduced, compared to fine-tuning the base LLM.

In practice, we expect that a large amount of such shared pLLMs will be available on a trusted cloud server, due to the following reasons. First, sharing pLLMs incurs the minimum risk of privacy leakage, because it is difficult to precisely recover the original training data from a trained model \cite{morris2023language,skapars2024slander}. Even if recovery is possible in some cases by model inversion \cite{morris2023language,skapars2024slander}, it remains unclear whether the extracted knowledge originates from pre-training data or the user's private data for fine-tuning. Hence, users are well motivated to share their pLLMs. Second, LLM personalization has been nowadays a common practice, which even non-professional users can easily operate by following the instructions or using the APIs given by the model publisher. For example, Anyscale's fine-tuning API \cite{anyscale} allows individuals to use a cloud server to personalize any open-sourced LLM with a single command. The large population of users in LLM personalization, hence, ensures that pLLMs fine-tuned by personal data with diverse characteristics would be available.


The effectiveness of LLM personalization, then, depends on appropriate selection of the cached pLLM. Intuitively, the selected pLLM should have been fine-tuned with data that is similar to the current user's personal data, which however, cannot be uploaded to the server for pLLM selection due to privacy restrictions. Alternatively, the user can download all the cached pLLMs and evaluate each pLLM with personal data \cite{peng2023check,salazar2019masked}. However, doing so produces high overheads in both communication (for transmitting pLLMs to the local device) and computation (for local search among pLLMs), especially when the number of cached pLLMs is large.

Instead, in this paper we present \emph{XPerT} (eXplainable Personalized Tuning), a new technique that ensures appropriate pLLM selection based on explainability about how the cached pLLM was fine-tuned. More specifically, the server computes the explanation of each cached pLLM as the drift of its output distribution from such distribution of the pre-trained base LLM, and transmits explanations of cached pLLMs to the current user, which then locally compares these explanations with its personal data for model selection. 
To further ensure the user's trust about on-device LLM personalization, such explanations of pLLMs in XPerT should be contain sufficient details in natural language to be directly understandable to human users. For example, for LLMs being personalized by user-generated texts in different language styles, which is currently the most common way of how LLMs are personalized \cite{zhang2024personalization}, we aim to explain the language styles of which each pLLM is personalized\footnote{We will also demonstrate that the similar explainability on different styles of how image generation models are personalized can also be enabled by XPerT.}.

To achieve such explainability, our basic idea is to leverage the strong natural language capabilities of larger LLMs to summarize the different styles in pLLMs' generative outputs. More specifically, we use another large LLM as the summarizer to encode the aforementioned distribution drift into an embedding vector, and further decompose this vector into a linear combination of sub-vectors that are orthogonal in the embedding space. Each sub-vector represents an aspect (e.g., one language style) of the distribution shift, with its coefficient in the linear combination indicating its relative importance, and can then be transformed into natural language by being converted back to the token space. Based on the linear representation property of LLMs \cite{mikolov2013linguistic,elhage2022toy,wang2024concept,nanda2023emergent}, if these sub-vectors are orthogonal in the embedding space, the aspects of the distribution shift they represent are also orthogonal, and such orthogonality ensures that different aspects of natural language explanations well complement each other with the minimum redundancy.



This approach also allows the model selection in XPerT to be done in a fully automated manner. More specifically, the user similarly uses the summarization LLM to apply the same decomposition to its local personal data, whose coefficients in the decomposition will be used to search for a cached pLLM with the most similar coefficients. Such numerical matching of coefficients minimizes both the computation and communication overhead of model selection.


To our best knowledge, XPerT is the first that enables LLM model selection for LLM personalization at the user's device, providing full explainability without acquiring the full models themselves. We implemented XPerT on multiple Android smartphone models, and evaluated its performance with text generation datasets and various LLM families including Llama3 \cite{dubey2024llama}, Qwen2 \cite{yang2024qwen2} and SmolLM \cite{allal2024SmolLM}. Our experiment results lead to the following conclusions:

\begin{itemize}
	\item XPerT ensures highly accurate model selection with correct explainability. Such accuracy of model selection can be as high as 96\%, with different LLM models being fine-tuned in different language styles.
    \item Based on such accurate model selection, XPerT largely reduces the computation costs of on-device LLM personalization, without affecting the downstream task performance. When the selected pLLM matches the user's personalization need, the wall-clock fine-tuning time can be reduced by 83\%.
    \item When the proper pLLM is used for on-device LLM personalization, XPerT also improves the data efficiency of such personalization by up to 51\%.
    \item XPerT is lightweight. Its communication and computation overheads of on-device model selection are 96.5\% lower than the existing methods of model selection.
\end{itemize}

\begin{figure}[ht]
	\centering
		\vspace{-0.2in}
	\includegraphics[width=1\columnwidth]{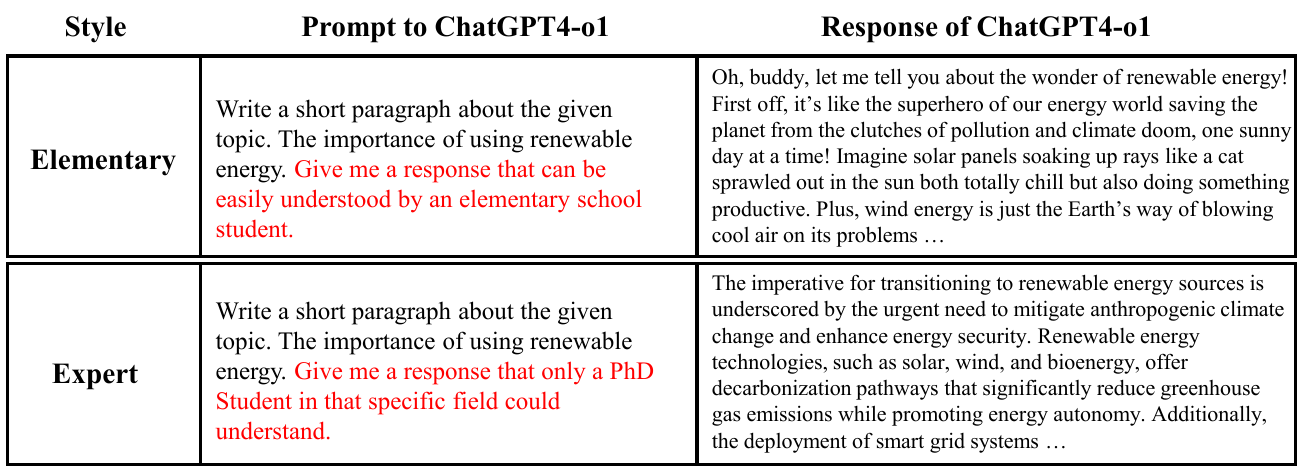} 	
	\vspace{-0.3in}
	\caption{An example of synthetic texts in different language styles. Texts in red are added to instruct ChatGPT4-o1 to follow different language styles when generating responses.}
	\vspace{-0.25in}
	\label{fig:synthetic_example}
\end{figure}

\begin{figure*}[ht]
	\centering
	\hspace{-0.2in}
	\subfigure[Llama3.2-1B \cite{dubey2024llama}] { 
		\includegraphics[width=0.305\textwidth]{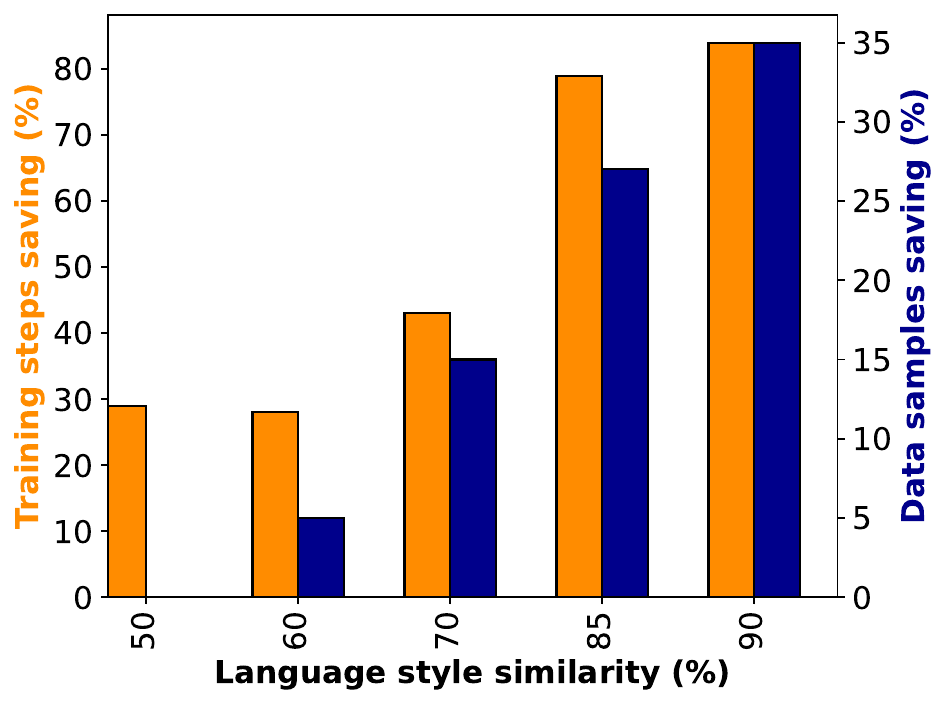}
		\label{fig:delay_impact}
	}
	\hspace{0.1in}
	\subfigure[Qwen2-0.5B \cite{yang2024qwen2}] { 
		\includegraphics[width=0.3\textwidth]{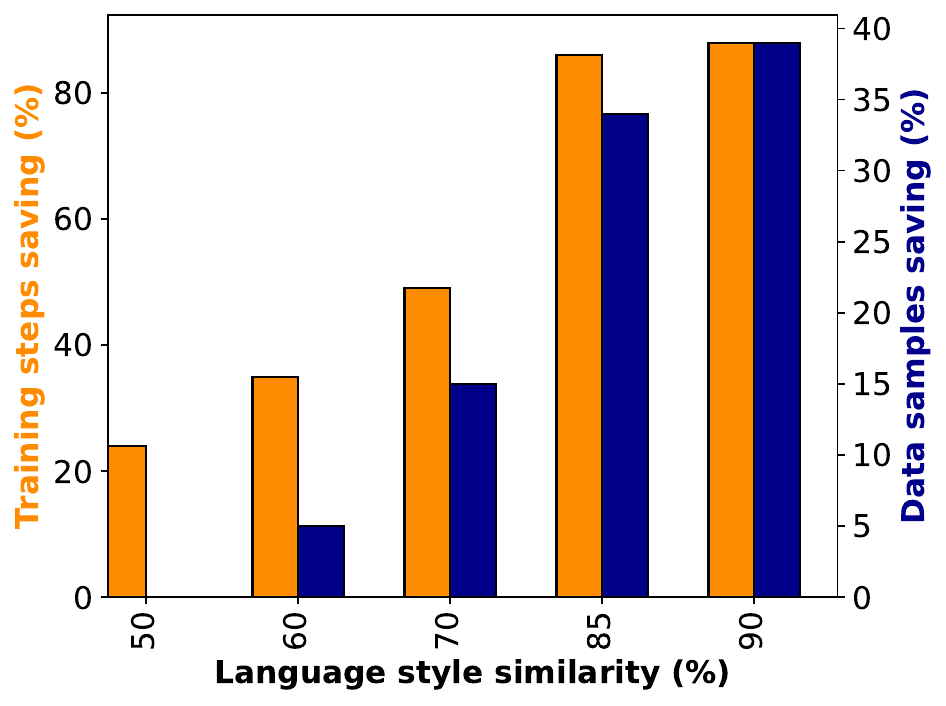}
		\label{fig:weighted_acc}
	}
	\hspace{0.1in}
	\subfigure[SmolLM-360M \cite{allal2024SmolLM}] { 
		\includegraphics[width=0.31\textwidth]{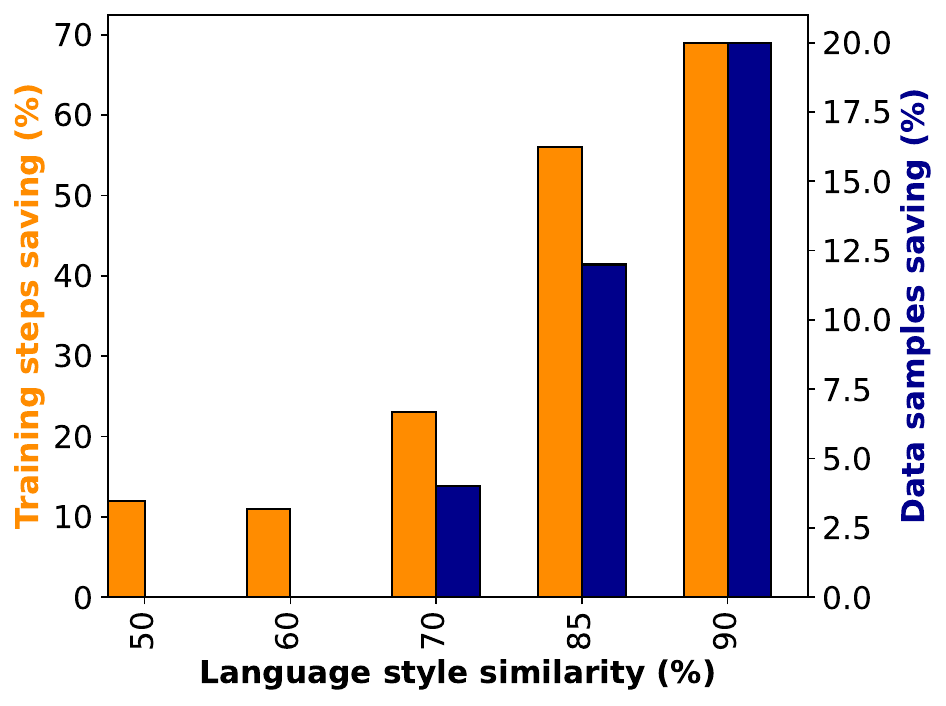}
		\label{fig:first_order_error}
	}
	\hspace{-0.1in}
	\vspace{-0.2in}
	\caption{Model selection reduces the computational cost and the amount of data required for LLM fine-tuning}
	\label{fig:prelim_reduction}
	\vspace{-0.15in}
\end{figure*}

\vspace{-0.05in}
\section{Background and Motivation}

\subsection{Benefits of Fine-tuning the pLLM}
Our design of XPerT builds on the following hypothesis of model selection: if the selected pLLM exhibits the similar output distribution as the distribution of current user's personal data, fine-tuning the selected pLLM with this personal data will incur lower computing costs and require less training data, compared to fine-tuning a pre-trained base LLM. 

To verify this hypothesis, we prompt ChatGPT4-o1 to synthesize text responses in different language styles to randomly sampled instructions from the Alpaca dataset \cite{alpaca}, and then use synthetic text data in different language styles to train different pLLMs. As shown by the example in Figure \ref{fig:synthetic_example}, we follow the same setting in \cite{jang2023personalized} to use 16 language styles, as different combinations of three aspects: 1) expertise (elementary/expert), 2) informativeness (concise/informative), and 3) style (friendly/unfriendly/sassy/sarcastic). Then, we simulate the current user's personal data as mixtures of synthetic text data in different language styles, and examine the computing costs of further fine-tuning the trained pLLMs using such personal data. Results in Figure \ref{fig:prelim_reduction} show that, when the personal data shares high similarity with the data used to train the pLLMs, using such personal data to further fine-tune the trained pLLMs requires up to 80\% fewer training steps to converge, and requires up to 35\% less training data to reach the same testing loss.



\vspace{-0.05in}
\subsection{Explainable Model Selection}

\begin{figure}[h]
	\centering
	\vspace{-0.05in}
	\includegraphics[width=0.8\columnwidth]{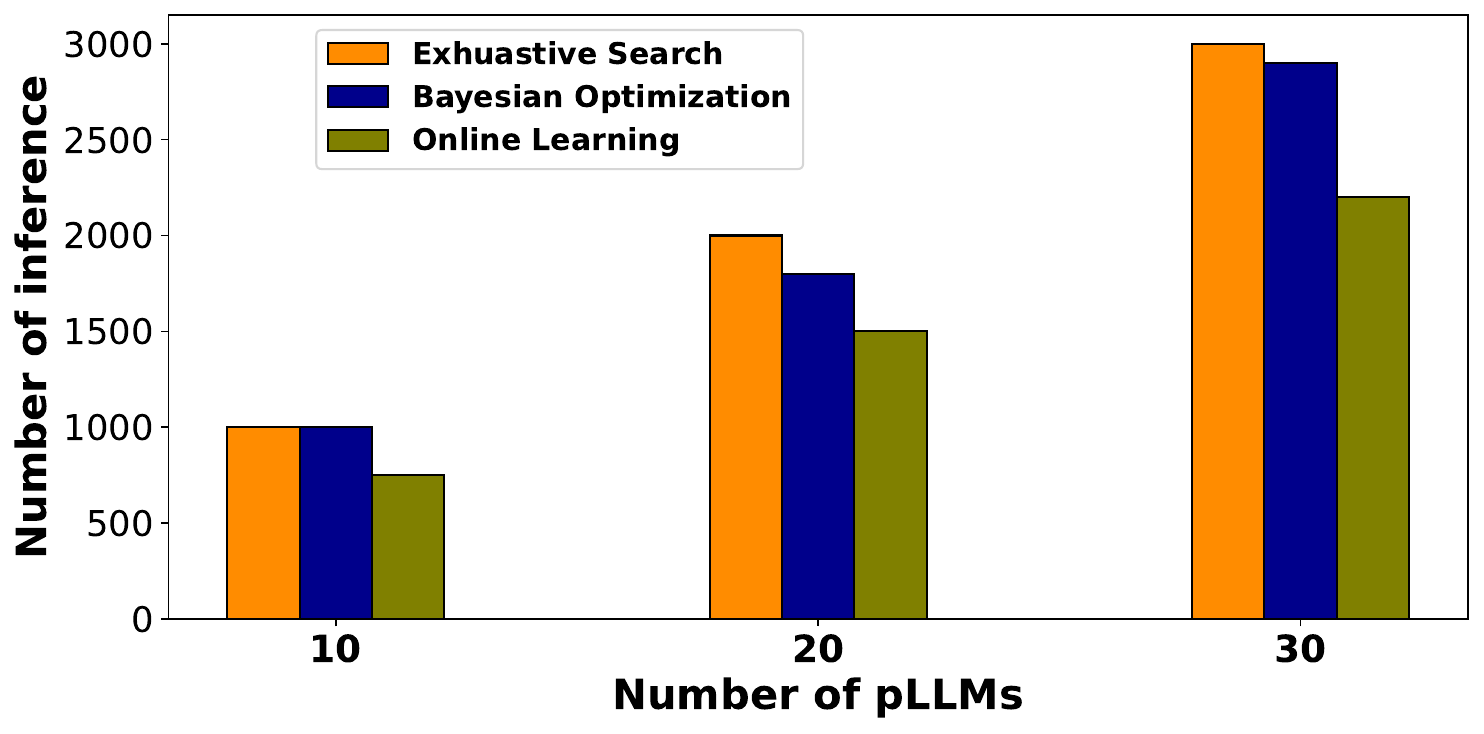} 	
	\vspace{-0.15in}
	\caption{Number of inferences needed for pLLM selection by evaluation with local personal data}
	\vspace{-0.1in}
	\label{fig:prelim_baseline}
\end{figure}

To ensure proper model selection, an intuitive approach is to download all the cached pLLMs and evaluate each of them using local personal data \cite{peng2023check,salazar2019masked}. However, as shown in Figure \ref{fig:prelim_baseline}, such evaluation is computationally expensive and does not scale well with the number of cached pLLMs, even with advanced search methods such as Bayesian optimization \cite{feurer2022auto,kotthoff2017auto} and online learning \cite{cutkosky2017online,foster2017parameter}, because pLLM inference is needed for each personal data sample. 

To avoid such expensive evaluation, the server can extract high-level information about how each cached pLLM is fine-tuned and the user selects the pLLM based on such extracted information. Early efforts calculate the importance scores of model features and area of expertise from the model's regions of competence \cite{cerqueira2017arbitrated,saadallah2022explainable}, but only apply to simple tree-based models \cite{jakobs2023explainable} or CNNs \cite{saadallah2021explainable}. For LLMs, an intuitive method is to use pre-trained text encoders \cite{radford2021learning,kim2024fine} to encode both the pLLMs' outputs and local personal data into feature vectors and make decisions of mode selection based on the similarity between feature vectors, but this method lacks explainability and hence cannot ensure correctness.

Instead, explainable model selection interprets such high-level information as the shift between output distributions of the cached pLLM and the base LLM. Some methods separately discover underlying patterns in these two output distributions using prototype learning \cite{van2023prototype,li2015mid,doersch2015makes} and then compare their differences, but involve iterative optimizations that are computationally expensive. One can also prompt another summarization LLM to directly compare the difference between pairs of samples from these two output distributions and summarize the difference into natural language \cite{zhong2022describing,singh2023explaining,zhong2023goal}, but such summarized difference could be biased and is difficult to be quantified for proper model selection.

Our approach in XPerT aims to address these limitations and is inspired by the recent technique of direct summarization \cite{zhong2022describing,singh2023explaining,zhong2023goal}. Instead of generating a single explanation of the cached pLLM, we extract such explanation as an embedding vector in the summarization LLM's embedding space and decompose it into a linear combination of orthogonal sub-vectors. In this way, each sub-vector represents one detailed aspect of explanation.

\vspace{-0.05in}
\section{Overview}

\begin{figure*}[h]
	\centering
	\vspace{-0.05in}
	\includegraphics[width=0.85\textwidth]{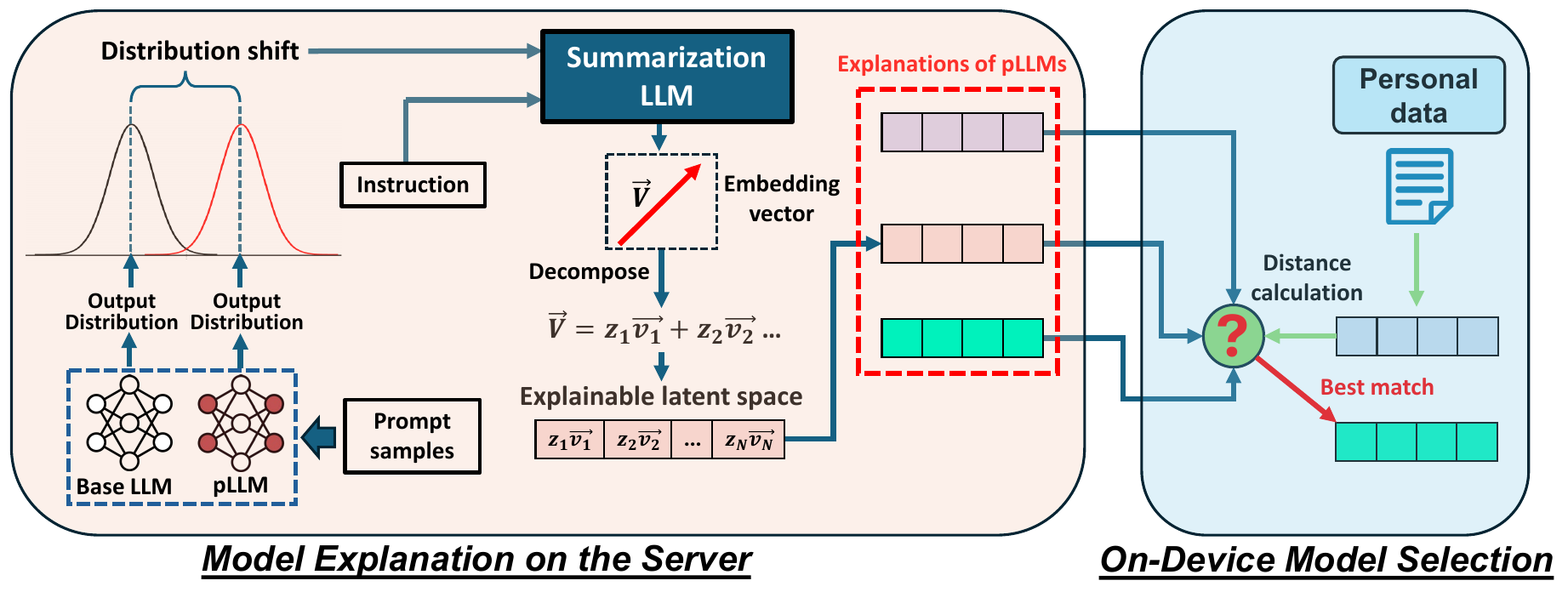} 	
	\vspace{-0.1in}
	\caption{Overview of XPerT design}
	\vspace{-0.1in}
	\label{fig:method_overview}
\end{figure*}

As shown in Figure \ref{fig:method_overview}, the basic design rationale of XPerT is to retain the computations needed for explaining pLLMs completely at the server, and only transmit the explanations of pLLMs to the user's device. As result, the user is able to select the server's cached pLLM for further on-device personalization by comparing the pLLMs' explanations with its local personal data, without either downloading all the pLLM models in full or exposing its personal data to the server. The user's local communication and computation overheads, hence, could both be minimized. 

\vspace{-0.05in}
\subsection{Model Explanation on the Server}
\label{subsec:overview_server}

Given the pre-trained base LLM and a pLLM cached at the server, our objective of model explanation at the server is to provide a natural language description about the difference between their output distributions in a quantitative manner.


\begin{figure}[ht]
	\centering
	\includegraphics[width=0.9\columnwidth]{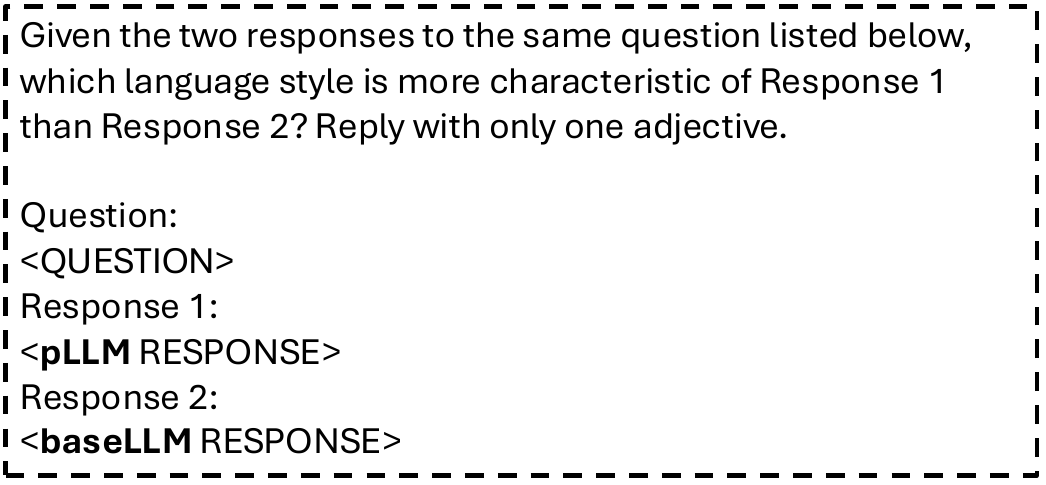} 	
	\vspace{-0.1in}
	\caption{The instruction for the summarization LLM}
	\vspace{-0.15in}
	\label{fig:prompt_template}
\end{figure}

To achieve this objective, we probe both the base LLM and pLLM with the same set of prompt samples, and then employ another summarization LLM to summarize the difference of their output distributions in language style, using the instruction shown in Figure \ref{fig:prompt_template}. To quantify such summarized difference, 
we first extract one embedding vector ($\vec{V}$) as the hidden states before the summarization LLM's output layer, and then decompose this embedding vector into a linear combination of orthogonal sub-vectors, such that 
\begin{equation*}
\vec{V}=\sum\nolimits_{i=1}^N z_i \vec{v}_i,
\end{equation*}
where $\vec{v}_i \cdot \vec{v}_j = 0$ for any $i \neq j$, so that the coefficients $\{z_i\}$ in the linear combination can be used as quantitative measures. Note that, since the summarization LLM is instructed to describe the output difference as only one single adjective word, this embedding vector is ensured to contain all the relevant information about the summarized output difference. In practice, to avoid possible errors and bias in probing the base LLM and pLLM, we shall use a sufficiently large and representative set of prompt samples. Details of ensuring such accurate and unbiased probing are in Section \ref{subsec:extraction_vector}.

Our approach to decomposing the embedding vector builds on the existing empirical studies \cite{mikolov2013linguistic,elhage2022toy,wang2024concept,nanda2023emergent}, which suggest that high-level concepts in a LLM are represented linearly in the LLM's embedding space. This linearity ensures that our decomposition is also linear, and we aim to let each sub-vector ($\vec{v}_i$) in such decomposition represent a simple concept (i.e., one language style), as one aspect of the complex aspect represented by the embedding vector (i.e., how the pLLM is personalized). The coefficients ($\{z_i\}$), then, indicate how much the pLLM was personalized in different language styles, and represent a coordinate $\mathbf{Z}$ in an explainable latent space $\mathbb{R}^n$ of the summarizing LLM. The difference between pLLMs can be quantitatively measured as the distances between their coordinates in this latent space. Such decomposition will be sequentially applied on each pLLM, to obtain a common set of sub-vectors for all cached pLLMs. More details are in Section \ref{subsec:decomposition_vector}.

\vspace{-0.05in}
\subsection{On-Device Model Selection}
\label{subsec:overview_device}

The server only transmits the common set of orthogonal sub-vectors of all cached pLLMs, as well as the coordinates of these pLLMs in the explainable latent space, to the user's device. The user uses the same method in Section \ref{subsec:overview_server} to map its personal data to the set of sub-vectors, and selects the pLLM whose coordinate is the closet to the coordinate of its personal data, i.e.,
\begin{equation*}
\arg \min\nolimits_i \texttt{Distance}(\mathbf{Z}_{pLLM_i}, \mathbf{Z}_{local}).
\end{equation*}

The user only downloads this selected pLLM from the server and uses it for on-device personalization with personal data. Details about such model selection are in Section \ref{subsec:model_partitioning}.

In some cases, even the best matching pLLM may still largely mismatch the user's personal data. In these cases, we instead aim to find a set of multiple pLLMs, which could better match the user's personal data if properly merged, i.e., 

\vspace{-0.1in}
\begin{equation}
\texttt{Distance}(\sum\nolimits_{i=1}^k \alpha_i \mathbf{Z}_i, \mathbf{Z}_{local}) < \tau,
\end{equation}
where $\mathbf{Z}_i$ and $\mathbf{Z}_{local}$ indicates the coordinate of the $i$-th pLLM and user's personal data, respectively, and $\tau$ is a threshold measuring the goodness of model selection. We use task arithmetic \cite{ilharco2022editing,ortiz2024task} for such model merging \cite{Survery_ModelMerging_2024}, and more details are in Section \ref{subsec:model_merging}.

\section{Model Explanation with Quantitative Measures}
\label{sec:model_explanation}

In this section, we provide technical details about extracting the embedding vector from a cached pLLM at the server and further decomposing this embedding vector into a linear combination of sub-vectors with quantitative measures.

\vspace{-0.05in}
\subsection{Extraction of the Embedding Vector}
\label{subsec:extraction_vector}
For each cached pLLM at the server, we probe both the pre-trained base LLM and the pLLM with the same set of prompt samples, and use the responses to extract the embedding vector from the summarization LLM's embedding space. The correctness of such embedding vector, then, depends on both the volume and contents of these prompt samples.

\begin{figure}[h]
	\centering
	\vspace{-0.05in}
	\includegraphics[width=0.85\columnwidth]{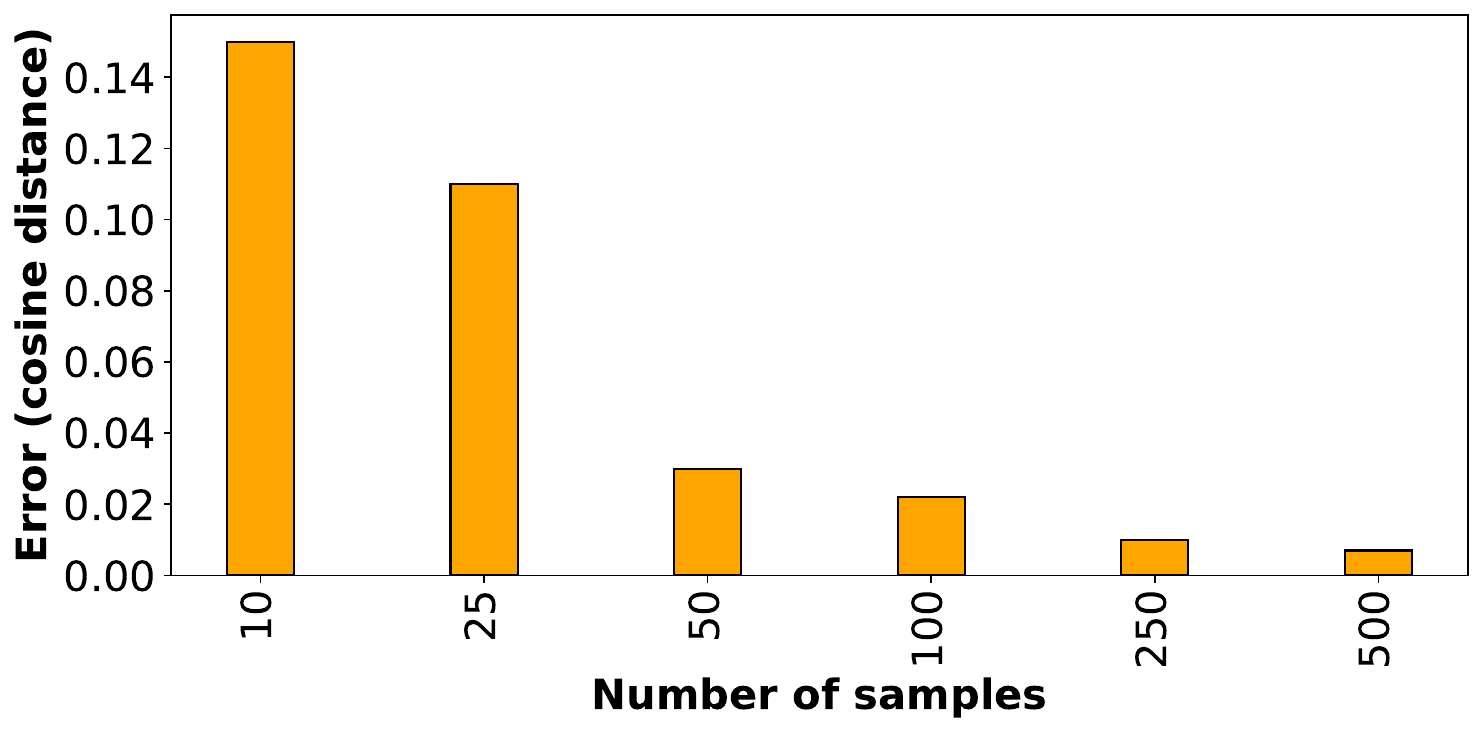} 	
	\vspace{-0.1in}
	\caption{The error of embedding vector with prompt numbers, measured by cosine distance}
	\vspace{-0.1in}
	\label{fig:sufficient_sample}
\end{figure}

To determine the required volume of prompt samples, we conducted experiments to examine how this volume correlates to the error of extracting the embedding vector, measured as the cosine distance from the embedding vector extracted with a sufficiently large volume of samples (e.g., 1,000). We use the instructions sampled from the Alpaca dataset \cite{alpaca} as prompt samples, and Llama-3.1-8B-Instruct \cite{dubey2024llama} as the summarization LLM. Results in Figure \ref{fig:sufficient_sample}, the error in the extracted embedding vector is negligible when the number of prompt examples reaches 50, and we will use this volume of prompt samples in the rest of this paper.

Intuitively, the set of prompt samples should also be representative to avoid any bias in probing. However, the way a pLLM is personalized, i.e., its language style, is an intrinsic characteristic of the model and applies to all the model's outputs. For example, if the given prompt is ``write a poem about \texttt{<subject>}'', the model's output will always tend to exhibit a more ``poetic'' style, no matter what specific subject in the prompt is. In this case, when we compute the output distributions of the pLLM and the base LLM, the possible bias, if any, will always present in both models' outputs and hence be canceled when we compute the embedding vector as the difference between these two distributions.

\begin{table}[ht]
	\centering
	\begin{tabular}{|c||c|c|c|}
		\hline
		\multirow{2}{*}{pLLM model}    & Llama3.2  &Qwen2 &SmolLM        \\ 
		& -1B & -0.5B & -360M \\ \hline \hline
		Alpaca vs. LegalBench & 0.042 &0.054&0.060\\ \hline
		LegalBench vs. SciQ &  0.063 &0.033&0.035 \\ \hline
		SciQ vs. Alpaca & 0.029 &0.041&0.032\\ \hline		
	\end{tabular}
	\vspace{0.05in}
	\caption{Differences of embedding vectors with prompt samples from datasets of different domains}	
	\label{tab:biased_sampling}
	\vspace{-0.25in}
\end{table}

To further verify this, we conducted experiments to evaluate the impact of prompt distribution on the embedding vector. We follow the same method and experimental settings in Section 2.1 to train a pLLM, compute its embedding vector using prompts samples from three datasets in different domains, namely Alpaca \cite{alpaca} with general instructions, LegalBench \cite{guha2023legalbench} with law-related questions, and SciQ with scientific questions \cite{welbl2017crowdsourcing}, and then measure the difference between these embedding vectors as the cosine distance. Results in Table \ref{tab:biased_sampling} show that, when using different LLM models as the pLLM and prompt samples from different domains, the embedding vectors calculated from the same pLLM remain consistent with small differences, demonstrating the minimal impact of the possible bias in prompt samples.

\begin{figure}[ht]
	\centering
	\vspace{-0.05in}
	\includegraphics[width=\columnwidth]{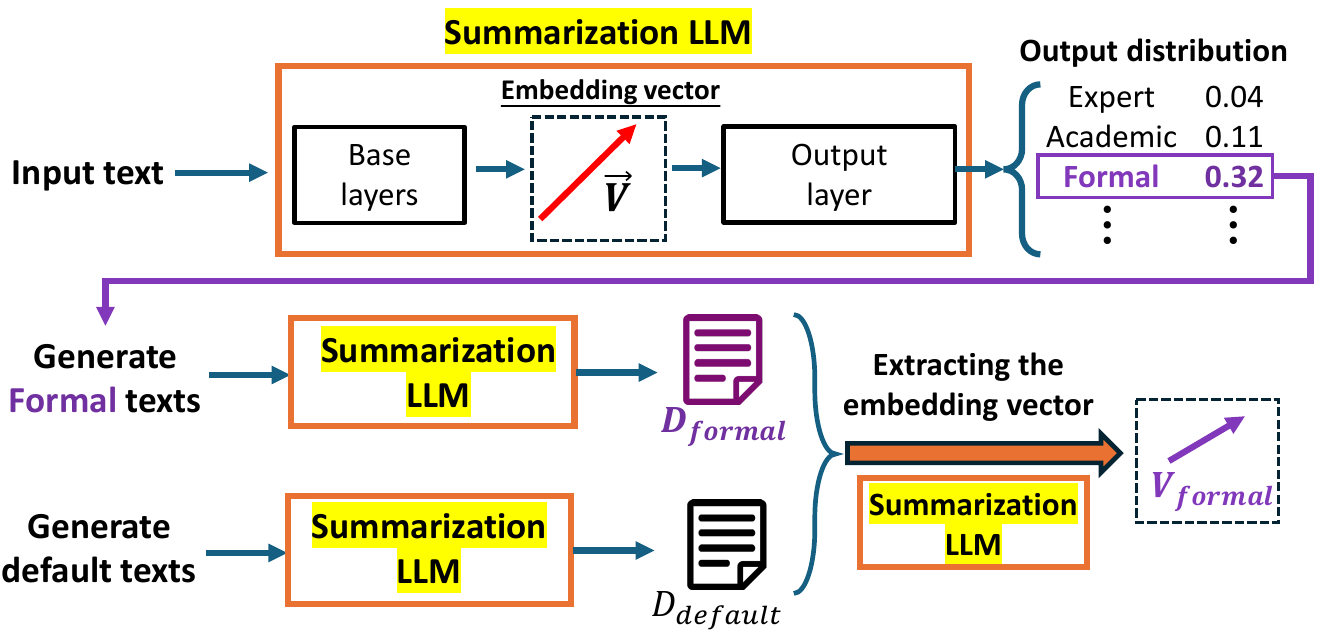} 	
	\vspace{-0.25in}
	\caption{Decomposing the pLLM's embedding vector into a linear combination of orthogonal sub-vectors}
	\vspace{-0.15in}
	\label{fig:sub_vector_finding}
\end{figure}

\subsection{Decomposing the Embedding Vector}
\label{subsec:decomposition_vector}


Our approach to decomposing the embedding vector into a linear combination of sub-vectors in the summarization LLM's embedding space, as shown in Figure \ref{fig:sub_vector_finding}, is to first check the summarization LLM's output distribution over its vocabulary, to obtain a list of candidate output words representing different language styles and their associated probabilities. For each candidate word, we instruct the summarization LLM to generate stylized texts in this language style, and use the similar method described in Section \ref{subsec:overview_server} to convert the distribution shift of these texts from those generated by the base LLM into an embedding vector, which then serves as the sub-vector representing this candidate word. The coefficient of this sub-vector is calculated by projecting the pLLM's embedding vector onto this sub-vector, such that
\begin{equation}
z_i=\vec{V} \cdot \vec{v}_i / \|\vec{v}_i\|.
\label{eq:coefficient}
\end{equation}

A sub-vector will only be used in the decomposition if it is orthogonal to all the existing sub-vectors. Since in practice it is hard to achieve 100\% orthogonality, we instead use a preset threshold in cosine distance to validate such orthogonality. We examine each candidate word in summarization LLM's output distribution following the descending order of their probabilities, until the linear combination of sub-vectors sufficiently approximates to the embedding vector, such that
\vspace{-0.1in}
	\begin{equation}
\lVert \vec{V}-\sum\nolimits_{i=1}^N z_i \vec{v}_i \rVert < \epsilon.
\label{eq:constraint_decomposition}
	\end{equation}
	

To facilitate model selection at the user's device, it is important that a common set of orthogonal sub-vectors is shared by all the cached pLLMs at the server, so that the user can match different pLLMs to its personal data in the same explainable latent space. To achieve so, we conduct the aforementioned decomposition sequentially over the cached pLLMs. For each pLLM, we first decompose its embedding vector using the set of sub-vectors calculated in the previous pLLM and examine if the constraint in Eq. (\ref{eq:constraint_decomposition}) is met. If not, we follow the same approach as depicted in Figure \ref{fig:sub_vector_finding} to add new sub-vectors with orthogonality to the existing sub-vectors, until the constraint in Eq. (\ref{eq:constraint_decomposition}) is met. In this way, the common set of orthogonal sub-vectors gradually expands until all pLLMs are processed. Such sequential process is described in Algorithm \ref{alg:sequential}.

\begin{algorithm}
	\caption{Finding a common set of sub-vectors ($\mathbf{S_v}$) shared by all the $N$ cached pLLMs at the server}
	\begin{algorithmic}[1]
		\State $\mathbf{S_v}\gets \{\}$
		\For{$i \gets$ 1 to $N$}
		\State Find candidate words for $pLLM_i$
		\For{each candidate $word_k$}
		\State $\vec{v}_i \gets $ Word-to-Vector Conversion ($word_k$) \textcolor{commentcolor}{// Generate the sub-vector as described in Figure \ref{fig:sub_vector_finding}}
		\If{$\vec{v}_i$ is orthogonal to all vectors in $\mathbf{S_v}$}
		\State $\mathbf{S_v} \gets \mathbf{S_v} \cup \{\vec{v}_i\}$
		\EndIf
		\State Decompose $\vec{V}_{pLLM_i}$ with $\mathbf{S_v}$ \textcolor{commentcolor}{// Coefficients are decided by Eq. (\ref{eq:coefficient})}
		\If{decomposition error $< \epsilon$ } \textcolor{commentcolor}{// The error is calculated by Eq. (\ref{eq:constraint_decomposition})}
		\State \textbf{break}
		\EndIf
		\EndFor
		\EndFor
	\end{algorithmic}
\label{alg:sequential}
\end{algorithm}

With this approach, the specific number of sub-vectors in this common set depends on how specifically pLLMs at the server are personalized. To explore such number in practice, we conducted experiments by using 0.1 as the threshold of orthogonality and $\epsilon=0.2\lVert \vec{V} \rVert$. We use Llama3.2-1B for the pLLMs and Llama3.1-8B-instruct as the summarization LLM, and follow the same method in Section 2.1 to fine-tune the pLLMs, such that each pLLM represents one different language style. Results in Table \ref{tab:vector_number} show that it needs multiple sub-vectors to fully represent one language style, and the number of orthogonal sub-vectors in the decomposition increases almost linearly with more pLLMs. Furthermore, the decomposition always needs to examine a much larger number of candidate words, as many words may share the similar meanings related to language styles.

\begin{table}[ht]
	\centering
	\begin{tabular}{|c||c|c|c|}
		\hline
		Number of pLLMs    & 4  &8 & 16       \\ \hline \hline
		Number of orthogonal sub-vectors & 12 & 19&37\\ \hline
		Number of candidate words examined & 92 & 145& 280\\ \hline		
	\end{tabular}
	\vspace{0.05in}
	\caption{Numbers of orthogonal sub-vectors and candidate words needed for the decomposition}	
	\label{tab:vector_number}
	\vspace{-0.3in}
\end{table}


\subsection{Computationally Efficient Sub-vector Generation}
\label{subsec:efficient_computation}

Since examining each candidate word as depicted in Figure \ref{fig:sub_vector_finding} requires multiple inferences in the summarization LLM to generate stylized texts and summarize these styles, the large number of candidate words to be examined, as shown in Table \ref{tab:vector_number}, suggest that generating sub-vectors is computationally expensive in practice. To improve the computational efficiency, 
we build on the strong natural language capability of the summarization LLM\footnote{The summarization LLM has much strong capability compared to on-device LLMs, due to its larger parameter sizes. For example, the Llama3.1-8B model outperforms its smaller counterpart, such as Llama3.2-1B model, by up to 3.2x on mainstream benchmarks \cite{nous}.}, and directly prompts the summarization LLM to generate the sub-vector from a candidate word, instead of delivering the knowledge about the candidate word via the stylized texts.

\begin{figure}[h]
	\centering
	\vspace{-0.1in}
	\includegraphics[width=0.85\columnwidth]{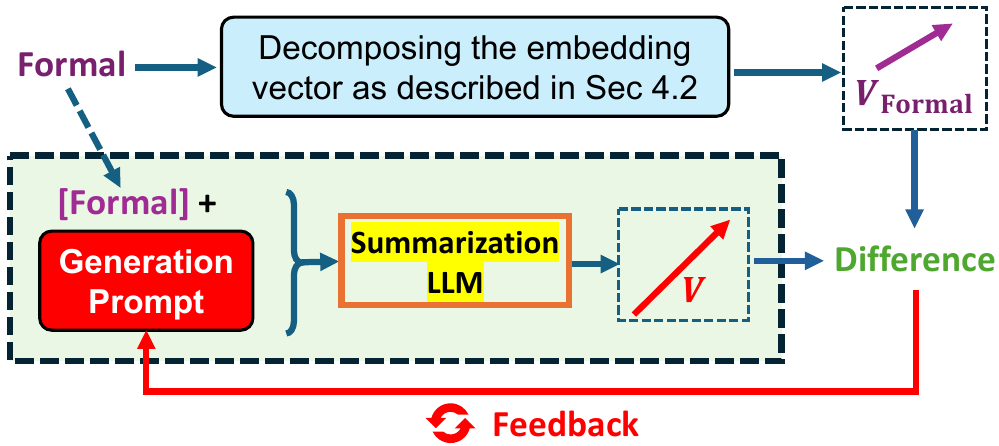} 	
	\vspace{-0.1in}
	\caption{Computationally efficient decomposition of the embedding vector into sub-vectors}
	\vspace{-0.1in}
	\label{fig:efficient_finding}
\end{figure}

To properly prompt the summarization LLM and hence minimize the error in the generated sub-vector, we design the prompt as [\texttt{prefix, candidate word, suffix}], where the \texttt{prefix} and \texttt{suffix} represent trainable tokens and are initialized with hand-crafted text. For example, the prompt can be initialized as ``use one adjective to summarize the meaning of \texttt{<candidate word>}, when it is used to describe a language style''. As shown in Figure \ref{fig:efficient_finding}, we first apply offline prompt tuning to optimize this prompt as trainable embeddings in summarization LLM's input space, and then use the optimized prompt for online sub-vector generation. We use the sub-vector generated by the method in Section \ref{subsec:decomposition_vector} as the ground truth, whose difference from the currently generated sub-vector is used as the loss in backpropagation.

To verify the accuracy of sub-vector generated by this method, we conducted experiments with different numbers of trainable tokens and measure the error of the generated sub-vector as its cosine distance from the aforementioned ground truth. As shown in Table \ref{tab:prompt_tuning_error}, when 200 samples of instruction-response pairs are used as training data in prompt tuning, such error can be sufficiently minimized. Further involving more training samples, on the other hand, produces only negligible improvement to the accuracy of sub-vector generation but adds extra computing overhead.


\begin{table}[h]
	\centering
	\vspace{-0.05in}
	\begin{tabular}{|c||c|c|c|c|c|}
		\hline
		\# of training samples    &50 &100 &200 & 300 & 400          \\ \hline \hline
		10 trainable tokens & 0.25 & 0.12 &0.07 &0.07 &0.08\\ \hline
		25 trainable tokens & 0.19& 0.09& 0.06 & 0.05&0.06\\ \hline
		50 trainable tokens & 0.29& 0.15&0.07 & 0.07&0.05 \\ \hline
		
	\end{tabular}
	\vspace{0.05in}
	\caption{The error of computing the sub-vector with prompt-tuned summarization LLM}	
	\label{tab:prompt_tuning_error}
	\vspace{-0.3in}
\end{table}

\section{On-Device Model Selection}
\label{sec:on_device_selection}
In this section, we describe details about model selection on the user's mobile device, using the explanations of pLLMs generated by the cloud server as described in Section \ref{sec:model_explanation}.


\subsection{Summarization LLM on Mobile Device}
\label{subsec:model_partitioning}
As described in Section \ref{subsec:overview_device}, the user's device needs to map its personal data to the common set of sub-vectors representing the server's cached pLLMs, to select the pLLM that best matches its personal data. However, such mapping requires inference of the summarization LLM, which has a large parameter size and may not fit to the memory constraint of mobile devices. For example, inference of a Llama3.1-8B model, even with INT8 precision, requires 10 to 12 GB spare memory, which is unavailable on most smartphones.

\begin{figure}[h]
	\centering
	\vspace{-0.05in}
	\includegraphics[width=0.65\columnwidth]{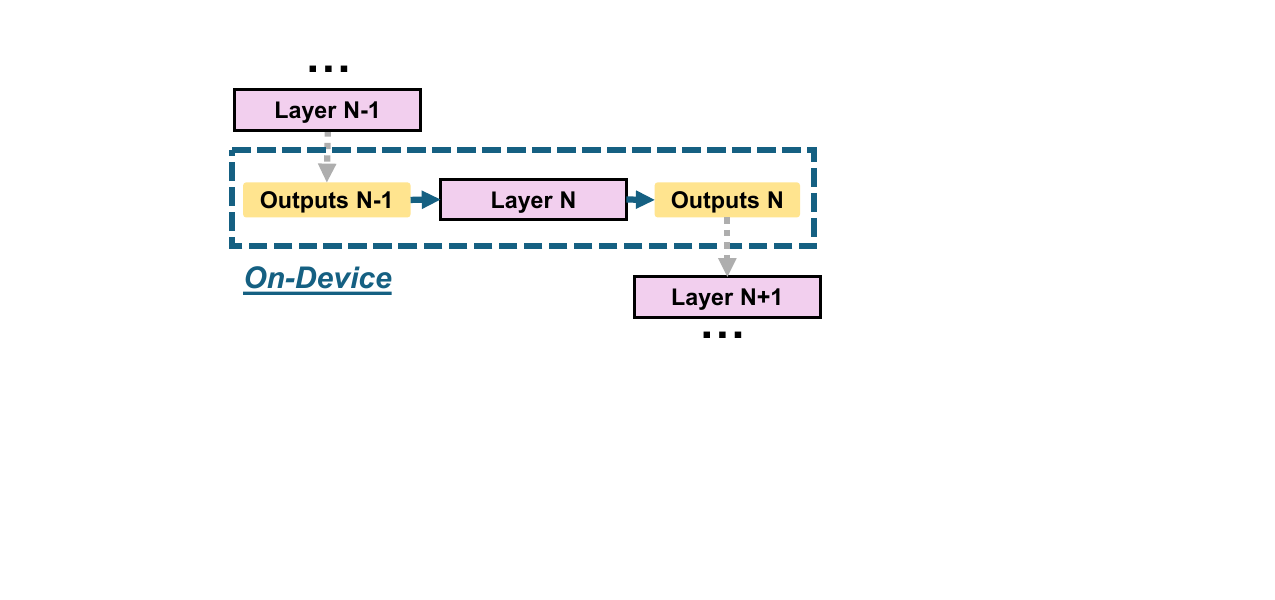} 	
	\vspace{-0.1in}
	\caption{Fitting the summarization LLM to the user device's memory constraint}
	\vspace{-0.2in}
	\label{fig:model_partition}
\end{figure}

To address this challenge, as shown in Figure \ref{fig:model_partition}, we can partition the summarization LLM and only perform inference of one layer at a time. The output of this layer is then cached in memory and used as the input to the next layer. In particular, since extracting the embedding vector, as described in Section \ref{subsec:overview_server}, essentially only involves predicting the next token, each layer of the summarization LLM only needs to be loaded and inferred once, and such loading and switching time between layers can hence be minimized.
 
To examine such loading and switching time in practice, we conducted preliminary experiments on various smartphone models, using Llama3.1-8B as the summarization LLM. When 50 prompt samples are used to prompt the summarization LLM as described in Section 3 and each prompt contains 200 tokens, Table \ref{tab:model_partition} show that the time needed for loading and switching different LLM layers is less than 2\% that of full LLM inference, and such switching will not change the generated sub-vectors at all. Note that, although the inference of summarization LLM itself takes a few minutes at the user's device to finish, such summarization of the user's personal data only needs to be done once, as long as the on-device data for LLM personalization remains unchanged. Meanwhile, the memory consumption of the summarization LLM's inference is reduced to 2.8GB, only 23\% of loading and executing the full LLM. 

\begin{table}[h]
	\centering
	\vspace{-0.05in}
	\begin{tabular}{|c||c|c|}
		\hline
		Device   & Inference time & Loading \& switching time     \\ \hline \hline
		OnePlus 12R & 10.1min & 9.87s \\ \hline
		Pixel 9 Pro & 14.2min & 12.24s  \\ \hline
	\end{tabular}
	\vspace{0.05in}
	\caption{Costs of on-device model loading and switching}	
	\label{tab:model_partition}
	\vspace{-0.25in}
\end{table}


\vspace{-0.1in}
\subsection{On-Device Model Merging}
\label{subsec:model_merging}
If the best matching pLLM still largely mismatches the user's personal data, we can opt to merge multiple pLLMs in different language styles to better resemble the user's local context. More specifically, for the coordinates of two pLLMs, namely $\mathbf{Z}_i$ and $\mathbf{Z}_j$, if we can find $\alpha_i$ and $\alpha_j$ such that 
\begin{equation}
\texttt{Distance}(\alpha_i \mathbf{Z}^i + \alpha_j \mathbf{Z}_j, \mathbf{Z}_{local}) < \tau,
\label{eq:model_merging1}
\end{equation}
it means that the personal data's language style is partially reflected by both of these two pLLMs. To merge these two pLLMs, one intuitive approach is knowledge distillation \cite{fukuda2017efficient}, where the teacher model is an ensemble of these two pLLMs and the student model will be the merged pLLM. However, this approach always involves expensive training efforts.

A more computationally efficient method is to apply task arithmetic \cite{ilharco2022editing,ortiz2024task}, which is an empirical technique for model merging \cite{Survery_ModelMerging_2024}, to steer the pLLM's language style by directly modifying its parameters using arithmetic operations. For pLLMs $i$ and $j$, we first compute their task vectors as 
\begin{equation*}
\tau_i=\theta_{pLLM_i}-\theta_{base}, \texttt{   } \tau_j=\theta_{pLLM_j}-\theta_{base},
\end{equation*}
where $\theta$ represents a LLM's parameter weights, and these two pLLMs can be merged by adding the task vectors to the base LLM:
\begin{equation}
\theta_{merged}=\theta_{base}+\tau_i+\tau_j.
\label{eq:model_merging2}
\end{equation}

Existing studies suggest to apply weights to the task vectors in Eq. (\ref{eq:model_merging2}) based on the importance of the respective tasks \cite{yang2024model} for better performance of the merged pLLM, but deciding such optimal weights involves iterative optimization \cite{zhang2024knowledge,liu2024checkpoint}, which is too expensive on mobile devices. Instead, an empirical solution is to use the coefficients, i.e., $\alpha_i$ and $\alpha_j$ in Eq. (\ref{eq:model_merging1}), as the weights in task arithmetic, because their magnitudes reflect the importance of the corresponding language styles in the explainable latent space.

Note that in model merging, since the task vectors are linearly appended onto the base LLM, the sub-vectors of the merged pLLM are just the union of sub-vectors of the original pLLMs used in merging. Hence, explainability of the merged pLLM remains unaffected.

\vspace{-0.05in}
\section{Implementation}
\label{sec:implementation}

\noindent \textbf{The cloud server.} We use a Lambda Labs server with 4 Nvidia H100 GPUs, to produce pLLMs by fine-tuning the base LLM with full-parameter training, using the Adam optimizer with a weight decay of 0.01. The learning rate is 1e-4, with a cosine learning rate scheduler and 10 warm-up steps. We use Llama3.1-8B-Instruct as the summarization LLM, and three pre-trained LLMs with smaller parameter sizes as the base LLM to be fine-tuned, including Llama3.2-1B \cite{dubey2024llama}, Qwen2.5-0.5B \cite{yang2024qwen2} and SmolLM-360M \cite{allal2024SmolLM}. This server will also be used for calculating the pLLM explanation.

\begin{figure}[h]
	\centering
	\vspace{-0.1in}
	\includegraphics[width=0.9\columnwidth]{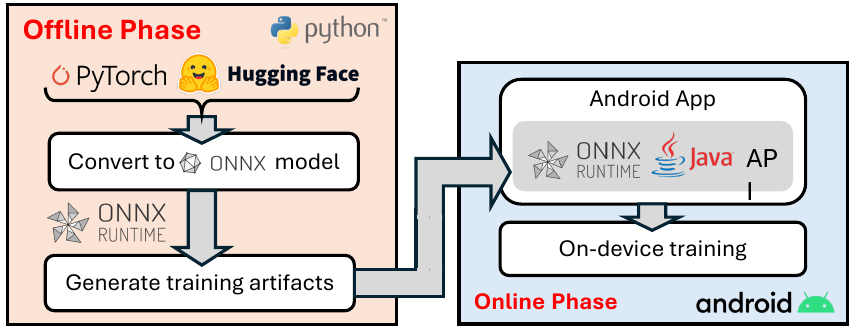} 	
	\vspace{-0.1in}
	\caption{Fine-tuning LLM on smartphones}
	\vspace{-0.15in}
	\label{fig:deployment}
\end{figure}

\noindent \textbf{On-device LLM fine-tuning.} Most of the existing mobile deep learning frameworks, including TensorFlow Lite \cite{tflite}, MNN \cite{alibaba2020mnn} and TVM \cite{chen2018tvm}, have limited support of fine-tuning LLMs on smartphones. Our implementation, as shown in Figure \ref{fig:delay_impact}, addresses this limitation by using the ONNX runtime framework \cite{onnx} on multiple Android-based smartphone models with different on-device computing power, including: 1) OnePlus 12R with Qualcomm Snapdragon 8 Gen 2 SoC, 2) Google Pixel 9 Pro with Google Tensor G4 SoC and 3) Google Pixel 7 with Google Tensor G2 SoC. On-device model selection will also be performed on these smartphones as described in Section \ref{subsec:overview_device}.

First, in the offline phase, we convert the HuggingFace pre-trained LLM (used as the base LLM in XPerT) to the ONNX format using the Optimum extension provided by the ONNX framework. During this process, we also applied LoRA \cite{hu2021lora} to reduce the computing cost. Afterwards, the ONNX Runtime Python API is used to generate the training artifacts with the base LLM and different training data, as the cached pLLMs. When a cached pLLM is selected for on-device personalization, we wrap the selected pLLM into an Android app on the smartphone, and further use the ONNX Runtime Java API to launch the Android app via Android Intent, with the training data provided through custom configuration files. On-device fine-tuning of the pLLM, then, is performed by this Android app as a background Android service.


\begin{table*}[ht]
\vspace{-0.05in}
	\begin{tabular}{|c||c|c|c|c||c|c|c|c||c|c|c|c||}
		
		\hline
		
		&\multicolumn{4}{c||}{\textbf{Llama-3.2-1B on One Plus 12R}} & \multicolumn{4}{c||}{\textbf{Qwen2-0.5B on Pixel 9 Pro}} &\multicolumn{4}{c||}{\textbf{SmolLM-360M on Pixel 7}} \\
		\hline
		\hline
		\textbf{Synthetic}& Acc  & FT-time &Energy &Data & Acc & FT-time &Energy &Data& Acc & FT-time &Energy &Data\\
		\hline
		\hline
		From scratch  &-&97.8min&15.7kJ&0\%&-&26.4min&6.2kJ&0\%&-&93.6min&18.0kJ&0\%\\
		30\% similarity  &25.0\%&92.4min&14.9kJ&4.6\%&28.6\%&23.1min&5.5kJ&2.4\%&17.8\%&89.4min&17.1kJ&3.6\%\\
		50\% similarity  &53.6\%&81.8min&13.3kJ&16.7\%&64.2\%&21.9min&5.2kJ&14.4\%&57.1\%&84.4min&16.7kJ&9.9\%\\
		70\% similarity&85.7\%&56.7min&9.0kJ&17.1\%&82.1\%&13.4min&3.1kJ&15.2\%&82.1\%&70.8min&13.6kJ&4.4\%\\
		80\% similarity&96.4\%&32.9min&5.3kJ&24.7\%&92.9\%&5.8min&1.4kJ&33.7\%&89.3\%&56.2min&10.9kJ&13.5\%\\
		90\% similarity&96.4\%&17.9min&2.8kJ&35.7\%&100\%&4.4min&1.0kJ&33.2\%&89.3\%&26.2min&5.1kJ&21.7\%\\
		\hline
		\hline
		
		\textbf{Real-world}& Acc & FT-time &Energy &Data & Acc & FT-time &Energy &Data& Acc & FT-time &Energy &Data\\
		\hline
		\hline
		From scratch &-&141.6min&23.5kJ&0&-&41.2min&9.40kJ&0&-&149.4min&26.5kJ&0\\
		30\% similarity  &16.4\%&124.2min&20.6kJ&7.2\%&16.9\%&38.5min&9.02kJ&3.7\%&8.3\%&140.7min&25.8kJ&5.2\%\\
		50\% similarity  &60.7\%&99.0min&16.6kJ&16.4\%&48.3\%&28.4min&6.58kJ&10.0\%&40.7\%&124.4min&24.1kJ&9.1\%\\
		70\% similarity&88.2\%&61.2min&10.1kJ&29.0\%&80.5\%&22.7min&4.34kJ&30.6\%&76.3\%&96.6min&17.2kJ&15.1\%\\
		80\% similarity&88.9\%&37.8min&6.27kJ&42.1\%&84.7\%&19.0min&19.0kJ&38.2\%&88.1\%&74.4min&12.9kJ&27.2\%\\
		90\% similarity &96.7\%&27.5min&5.29kJ&47.4\%&92.4\%&7.1min&1.61kJ&51.4\%&92.8\%&56.7min&10.7kJ&28.3\%\\
		\hline
	\end{tabular}
	\vspace{0.05in}
	\caption{XPerT's performance of on-device LLM personalization with different levels of similarity between pLLMs and on-device personal data. Performance is measured by model selection accuracy (`Acc'), wall-clock time (`FT-time') and energy consumption (`Energy') of on-device fine-tuning, and the improvement of data efficiency (`Data')}  
	\label{tab:finetune_saveing}
	\vspace{-0.3in}
\end{table*}

\vspace{-0.05in}
\section{Performance Evaluation}

In our evaluations, 
we use the following two text generation datasets with diverse language styles, for fine-tuning the base LLM into pLLM and on-device personalization of the selected pLLM:

\emph{1) Synthetic Data:} We prompt ChatGPT4-o1 to generate texts with 28 different language styles in three dimensions: 1) Expertise (elementary / expert), 2) Informativeness (concise / informative), and 3) Style (friendly/ unfriendly/ sassy/ sarcastic / persuasive / neutral / poetic), by sampling instructions from the Alpaca dataset \cite{alpaca} and applying them to the prompt in Figure \ref{fig:synthetic_example}, which asks for answers in a specific language style. For each language style, the dataset contains 52,000 question-answer (QA) pairs, and ChatGPT4-o1's default outputs without any language style are also included.

\emph{2) Real-World Data:} We combine three real-world datasets for more language styles, including 1) CDS (Corpus of Diverse Styles) \cite{style20}, containing 15 million sentences in styles of poetry, lyrics, tweets and Shakespeare; 2) Gutenberg3 \cite{gutenberg3}, containing extracts from 629 fiction novels in three styles, i.e., fantasy, romance, and sci-fi; and 3) ScientificPapers \cite{Cohan_2018}, containing two sets of long and structured documents obtained from the ArXiv and PubMed repositories as the academic style. We also mix data in two different styles to create intermediate styles, expanding the total to 25.

The personal data used for on-device LLM personalization and the training data for fine-tuning a pLLM will both contain only one language style, and hence the number of pLLMs cached at the server equals to the number of language styles in the datasets listed above. However, to mimic various levels of similarity between the on-device personal data and pLLMs, the training data for fine-tuning the pLLMs will also contain default samples without any style, and we then measure such similarity as the percentage of stylized samples in the pLLM's training data that have the same language style as the on-device personal data. All experiment results are then averaged over personal data with all language styles.

We compare XPerT with the existing methods of model selection. In particular, our scenario of model selection is related to the existing methods in AutoML \cite{feurer2022auto} and hyperparameter optimization \cite{heuillet2021sequential}, which are used as the baselines and listed below. 

\vspace{-0.05in}
\begin{itemize}
	\item \textbf{Exhaustive Search} \cite{peng2023check}, which evaluates each pLLM's output with the personal data and selects the best one.
	\item \textbf{Bayesian Optimization} \cite{feurer2022auto,kotthoff2017auto}, which treats the selection of pLLM as a new hyperparameter to be optimized by Bayesian optimization. 
	\item \textbf{HyperBand} \cite{falkner2018bohb,li2018hyperband} builds on the bandit principle to identify the best hyperparameter configurations and uses early stopping to eliminate bad ones.
\end{itemize}
\vspace{-0.05in}

Note that, XPerT is orthogonal to other parameter-efficient fine-tuning methods such as quantization \cite{egashira2024exploiting} and LoRA \cite{hu2021lora}, which can be applied to on-device personalization of the selected pLLM. These methods, hence, are not directly comparable with XPerT. Instead, LoRA has been applied to all our experiments of on-device LLM personalization.

\begin{table*}[ht]
	\centering
	\begin{tabular}{|c||c|c|c|c||c|c|c|c||c|c|c|c||}
		
		\hline
		
		&\multicolumn{3}{c||}{\textbf{Llama-3.2 1B on One Plus 12R}} & \multicolumn{3}{c||}{\textbf{Qwen2-0.5B on Pixel 9 Pro}} &\multicolumn{3}{c||}{\textbf{SmolLM-360M on Pixel 7}} \\
		\hline
		\hline
		\textbf{Synthetic}& BLEU& ROUGE-1 &ROUGE-L & BLEU& ROUGE-1 &ROUGE-L& BLEU& ROUGE-1 &ROUGE-L\\
		\hline
		\hline
		From scratch  &0.13&0.32&0.23&0.14&0.33&0.25&0.09&0.35&0.23\\
		30\% similarity&0.13&0.33&0.21&0.15&0.33&0.21&0.09&0.32&0.22\\
		70\% similarity   &0.12&0.33&0.21&0.13&0.32&0.24&0.10&0.34&0.21\\
		90\% similarity   &0.15&0.33&0.22&0.13&0.32&0.25&0.11&0.35&0.23\\
		\hline
		\hline
		\textbf{Real-world}& BLEU& ROUGE-1 &ROUGE-L & BLEU& ROUGE-1 &ROUGE-L& BLEU& ROUGE-1 &ROUGE-L\\
		\hline
		\hline
		From scratch&0.18&0.39&0.30&0.07&0.31&0.27&0.06&0.29&0.26\\
		30\% similarity &0.18&0.39&0.30&0.07&0.31&0.27&0.06&0.29&0.26\\
		70\% similarity &0.19&0.39&0.28&0.07&0.31&0.25&0.06&0.28&0.27\\
		90\% similarity &0.20&0.38&0.29&0.07&0.31&0.28&0.06&0.28&0.29\\
		\hline
		
	\end{tabular}
	\vspace{0.05in}
	\caption{XPerT's performance in text generation task, when using personalized LLM fine-tuned from the selected pLLM and from scratch, measured by the BLEU and ROUGE scores}  
	\label{tab:generation_quality}
	\vspace{-0.2in}
\end{table*}

\begin{table*}[ht]
	\centering
	\begin{tabular}{|c||c|c|c||c|c|c||c|c|c||}
		\hline
		&\multicolumn{3}{c||}{\textbf{Llama-3.2-1B on One Plus 12R}} & \multicolumn{3}{c||}{\textbf{Qwen2-0.5B on Pixel 9 Pro}} &\multicolumn{3}{c||}{\textbf{SmolLM-360M on Pixel 7}} \\
		\hline
		\hline
		\textbf{Synthetic}& $t_{comm}$ & $t_{comp}$  &Power & $t_{comm}$ & $t_{comp}$ &Power& $t_{comm}$ & $t_{comp}$  &Power\\
		\hline
		\hline
		Exhaustive Search  &1.26h&48.3min&13.1kJ&29.9min&19.6min&4.51kJ&23.8min&12.7min&1.54kJ\\
		Bayesian Optimization &1.26h&45.5min&12.1kJ&27.8min&17.3min&4.00kJ&22.9min&12.1min&1.46kJ\\
		HyperBand  &1.26h&40.0min&10.3kJ&29.9min&16.2min&3.76kJ&23.8min&10.6min&1.29kJ\\
		\textbf{XPerT}&2.7min&13.8min&3.7kJ&1.07min&10.8min&2.50kJ&0.85min&10.1min&1.22kJ\\
		\hline
		\hline
		
		\textbf{Real-world}& $t_{comm}$ & $t_{comp}$ &Power & $t_{comm}$ & $t_{comp}$ &Power& $t_{comm}$ & $t_{comp}$ &Power\\
		\hline
		\hline
		Exhaustive Search &1.12h&42.8min&11.4kJ&26.7min&16.2min&3.75kJ&21.2min&11.3min&1.54
		kJ\\
		Bayesian Optimization  &1.03h&39.3min&10.7kJ&24.6min&15.6min&3.75kJ&20.4min&10.7min&1.37kJ\\
		HyperBand  &1.12h&35.9min&9.6kJ&26.7min&14.4min&3.25kJ&21.2min&10.3min&1.12kJ\\
		\textbf{XPerT}&2.7min&13.8min&3.7kJ&1.07min&10.8min&2.50kJ&0.85min&10.1min&1.22kJ\\
		\hline		
	\end{tabular}
	\vspace{0.05in}
	\caption{Communication cost ($t_{comm}$), computation cost ($t_{comp}$) and power consumption of on-device model selection}  
	\label{tab:selection_cost}
	\vspace{-0.3in}
\end{table*}

\vspace{-0.05in}
\subsection{Performance of On-Device LLM Personalization}

We first evaluated XPerT's performance of on-device LLM personalization in the following three aspects:
\begin{itemize}
	\item \textbf{Accuracy of model selection} with various levels of similarity between the pLLMs and the on-device personal data, and a model selection is considered as accurate if the stylistic samples in the selected pLLM's training data shared the same language style with the on-device personal data.
	\item \textbf{On-device LLM personalization costs}, measured by the percentage of wall-clock time reduction (FT-time) and the amount of energy consumed (Energy).
	\item \textbf{Data efficiency}, measured by how much fewer data samples are required to achieve the same testing loss in on-device LLM fine-tuning.
\end{itemize} 

As shown in Table \ref{tab:finetune_saveing}, XPerT can ensure highly accurate model selection when the similarity between pLLMs and on-device personal data reaches 70\%, and can retain such accuracy at 60-65\% even if the similarity drops to 50\%. Based on such accurate model selection, XPerT can significantly reduce the cost of on-device LLM personalization. When the similarity is 70\%, XPerT can reduce the computing time needed for on-device LLM personalization by 55\% and the energy consumption by 57\%. When the similarity reaches 90\%, such reduction will further increase to 82\% and 83\%, respectively. On the other hand, even if the similarity is only 30\%, we can still reduce such cost by up to 15\%.

Meanwhile, Table \ref{tab:finetune_saveing} shows that fine-tuning the selected pLLM requires up to 51\% fewer data samples for on-device LLM personalization. Such improvement of data efficiency is important for mobile devices, where accumulating large amounts of personal data is usually very time consuming.

We also evaluated whether on-device personalization using the selected pLLMs would affect the personalized LLM's performance in text generation tasks. Since the cross-entropy loss may not align well with real-world evaluation for text generation tasks \cite{guo2023evaluating}, we instead used generation-based metrics including the BLEU \cite{Papineni02bleu:a} and ROUGE \cite{lin2004rouge} scores, which assess the quality of text generation by comparing the N-gram matches between generated texts and the ground truth. As shown in Table \ref{tab:generation_quality}, even with 30\% similarity, personalization using the selected pLLMs can achieve almost identical performance in downstream tasks, compared to fine-tuning the base LLM from scratch using the same training data.

\begin{figure*}[h]
	\centering
	\hspace{-0.2in}
	\subfigure[Communication Cost] { 
		\includegraphics[width=0.305\textwidth]{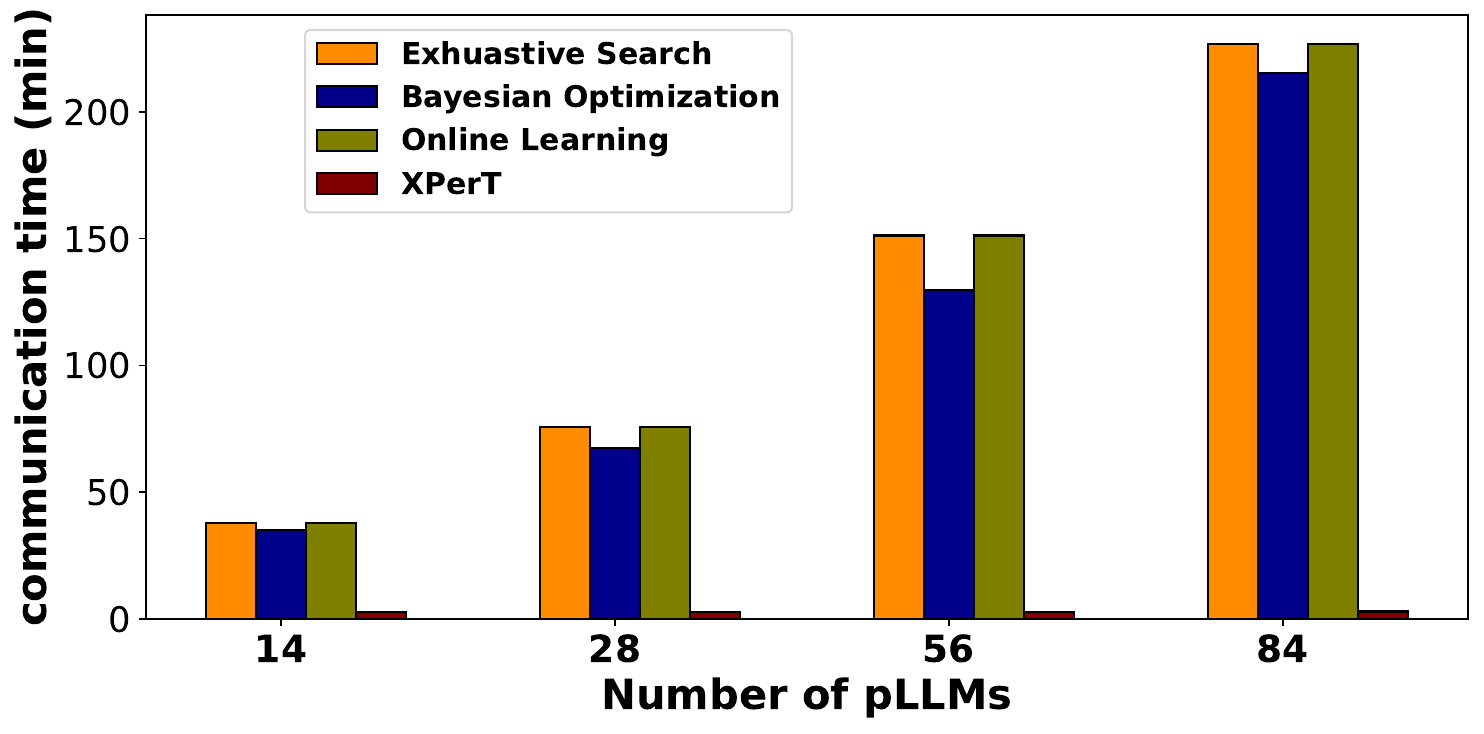}
		\label{fig:scalibility_comm}
	}
	\hspace{0.1in}
	\subfigure[Computing Cost] { 
		\includegraphics[width=0.3\textwidth]{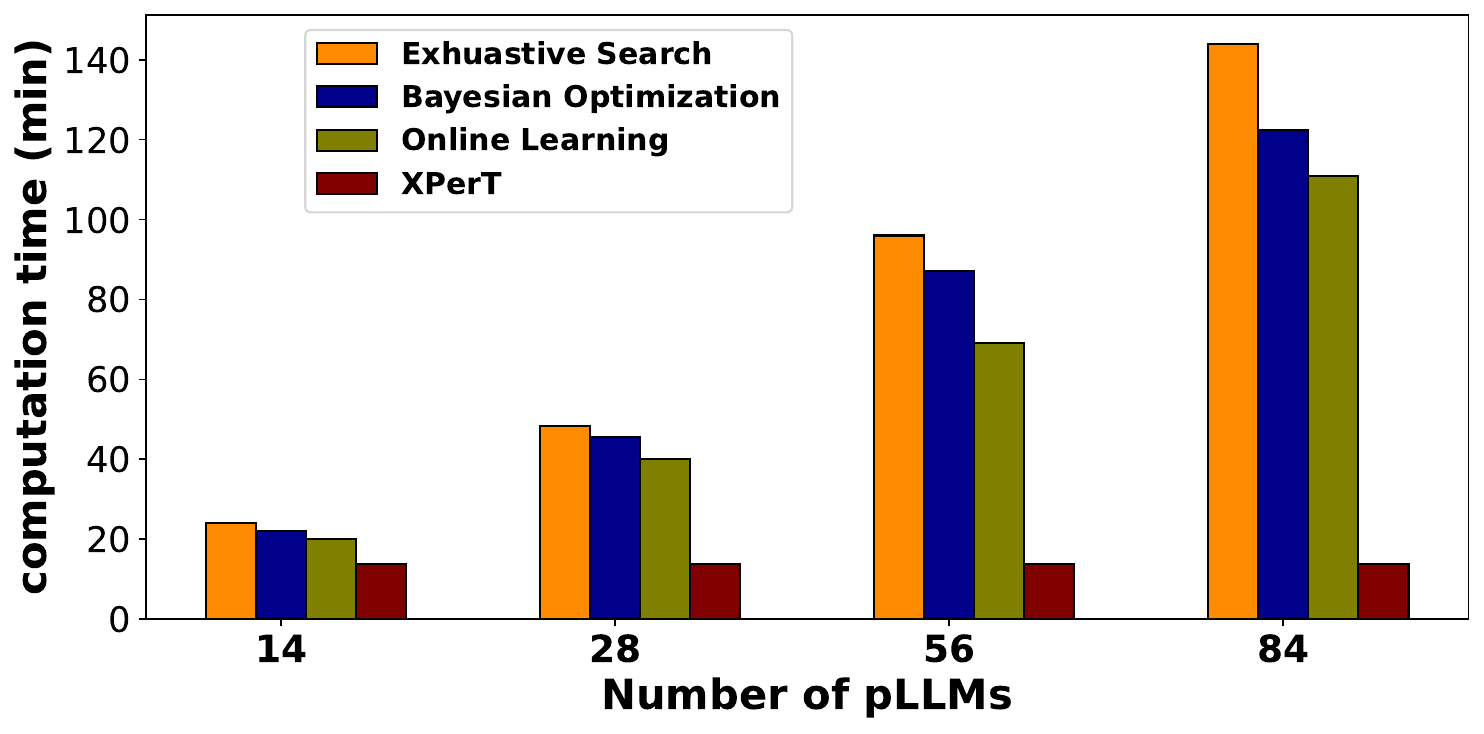}
		\label{fig:scalibility_comp}
	}
	\hspace{0.1in}
	\subfigure[Power Consumption] { 
		\includegraphics[width=0.31\textwidth]{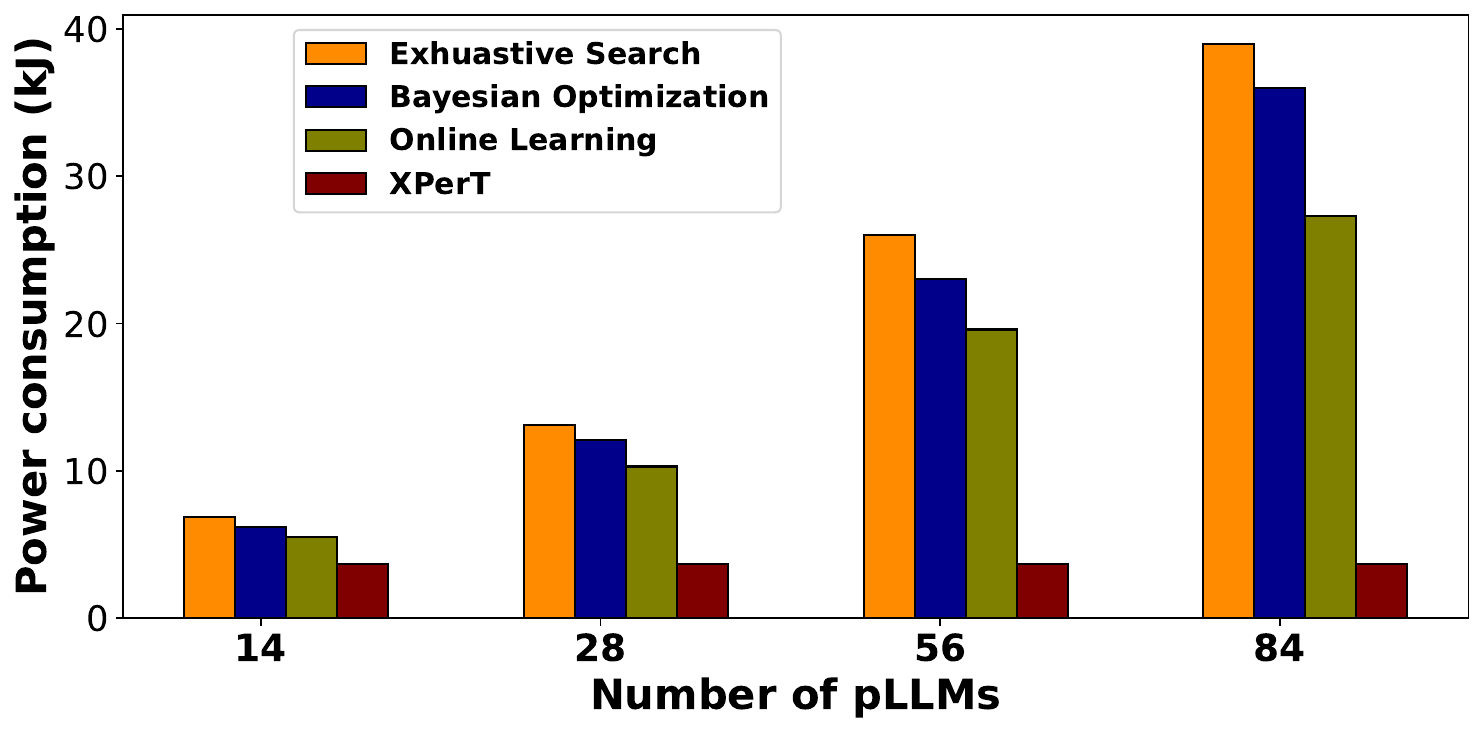}
		\label{fig:scalibility_power}
	}
	\hspace{-0.1in}
	\vspace{-0.2in}
	\caption{Scalability of XPerT's on-device model selection with different numbers of cached pLLMs on the server}
	\label{fig:scalibility}
	\vspace{-0.15in}
\end{figure*}

\vspace{-0.05in}
\subsection{Cost of On-Device Model Selection}

In XPerT, selecting the proper pLLM for on-device personalization is not for free, and instead incurs both communication and computation overhead to the user's mobile device. We investigated such model selection cost in XPerT and compared such cost with those in the existing methods of model selection. More specifically, the communication cost ($t_{comm}$) covers the time of downloading all the necessary data, models and model explanations from the cloud server, and the computing cost ($t_{comp}$) represents the wall-clock time for all the computations of model selection on smartphone devices. For fair comparisons, we use the same network bandwidth in all baselines and do not take the concurrent network traffic into account, because pLLMs can be transmitted only when the mobile device is idle to minimize the impact of such concurrent traffic. We also measured the end-to-end power consumption of on-device model selection, including data downloading, data loading, summarization LLM's inference, calculation of sub-vectors and coordinate search for the best pLLM. 


As shown in Table \ref{tab:selection_cost}, compared to baseline methods, XPerT significantly reduces both the communication and computation costs on the user's smartphone device, by up to 96.5\% and 71.4\%, respectively. For communication cost, XPerT avoids downloading any of the cached pLLM itself from the server, but instead only needs to transmit the set of extracted sub-vectors shared by all pLLMs and the pLLMs' coefficients in the corresponding explainable latent space. The computing cost, on the other hand, depend on the size of pLLM. For relatively large models such as the LLaMA3.2 1B, the computing cost of Exhaustive Search is approximately 3.5x higher than that of XPerT, because it needs to evaluate all the pLLMs with the on-device personal data via inference. Similarly, Bayesian Optimization achieves negligible savings in the computing cost, because the indices of pLLMs are randomly one-hot encoded and users lack information about how these pLLMs are fine-tuned. HyperBand can reduce the computing cost by 15-20\% via early stopping, but such saving is limited because a sufficient number of samples still should be evaluated to ensure accurate model selection.

XPerT's savings of computing costs become smaller when the model size is smaller, because the computing cost of summarization LLM inference becomes dominant. Nevertheless, such savings can still be up to 25\% with the synthetic dataset. The explainability of model selection in XPerT, further allows users to use the cloud server's cached pLLMs for on-device personalization with trust.


We also evaluated how the model selection cost varies with the number of cached pLLMs at the server. Results in Figure \ref{fig:scalibility} show that, while such costs in baseline methods linearly increase with the number of pLLMs, XPerT can retain a constantly low level of such cost, because all the pLLMs share a common set of sub-vectors that is used for on-device model selection. This scalability is important in many real-world scenarios that involve a large population of mobile users with much diverse language styles.

\begin{figure}[h]
	\centering
	\vspace{-0.1in}
	\includegraphics[width=0.7\columnwidth]{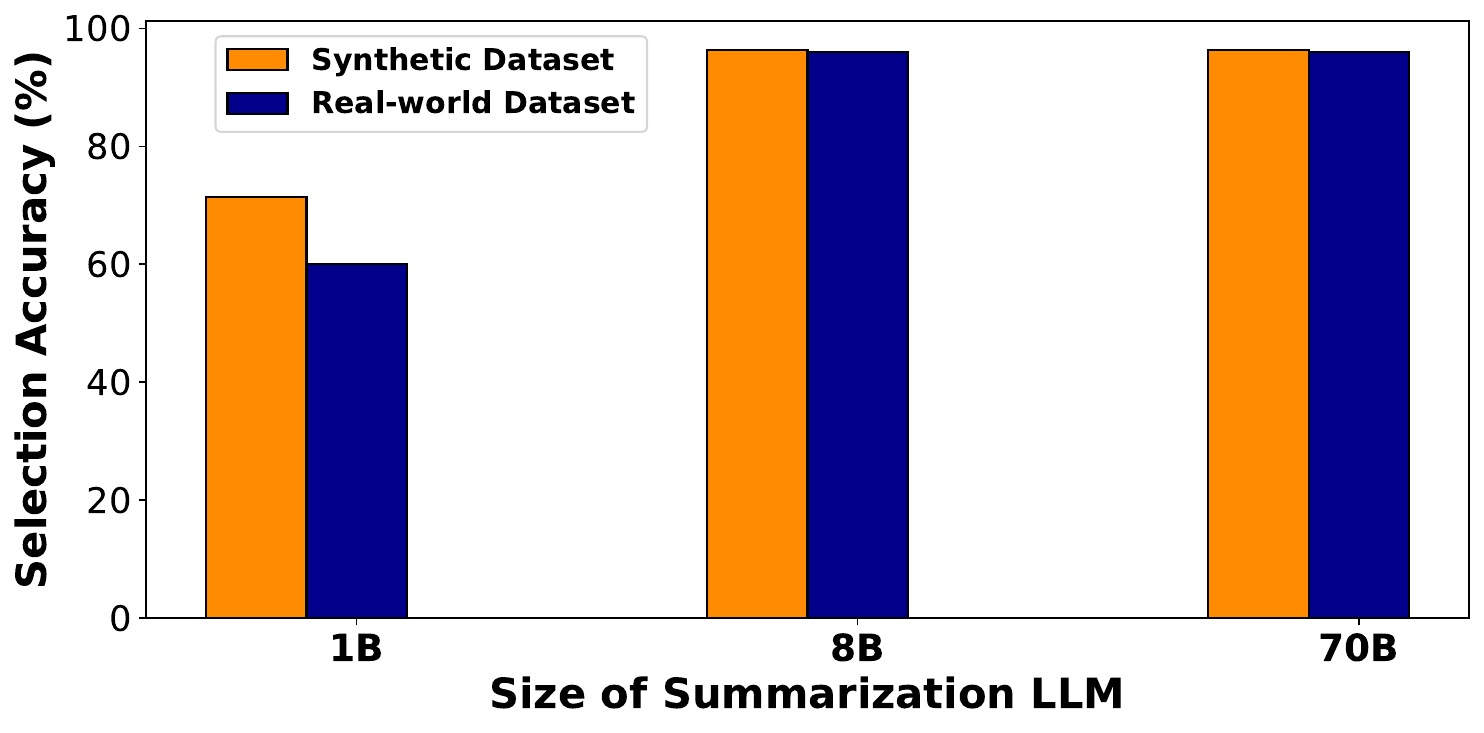} 	
	\vspace{-0.15in}
	\caption{Accuracy of model selection with variants of Llama3.1/3.2 models as summarization LLM}
	\vspace{-0.2in}
	\label{fig:sllm_size}
\end{figure}

\subsection{Using Different Summarization LLMs}

The performance of XPerT highly depends on the natural language capabilities of the summarization LLM, which generally relates to the parameter size of the summarization LLM. We evaluate XPerT's selection accuracy using different summarization LLMs, in a simulated environment on a server equipped with 4 NVidia H100 GPUs. As shown in Figure \ref{fig:sllm_size}, when we use the Llama3.2-1B model as the summarization LLM, the accuracy of XPerT's model selection is noticeably affected, due to the model's limited capabilities of summarizing the language styles and differences in pLLMs' output. However, when the model's parameter size increases to 8B, such impact on the model selection accuracy can be effectively eliminated. Further using a larger summarization LLM, such as the Llama3.1-70B-instruct model, brings little improvement on the model selection accuracy, but would be very expensive and infeasible on smartphones.

\vspace{-0.05in}
\subsection{The Impact of Empirical Thresholds}

As described in Section \ref{subsec:decomposition_vector}, we used two empirical thresholds when decomposing the embedding vector of a pLLM into sub-vectors: one threshold for deciding orthogonality between sub-vectors and another threshold for the linear combination of sub-vectors to approximate the embedding vector. We examined the performance of XPerT with different values of these two thresholds. Results in Tables \ref{tab:orthog_thr} and \ref{tab:epsilon} show that, using low values in these thresholds result in some redundant calculations of sub-vectors and impair the computing efficiency. In contrast, higher values of these thresholds may miss important sub-vectors and affect the accuracy of model selection. Therefore, in our experiments, we set the orthogonality threshold to 0.1 and the decomposition threshold to 0.2$\cdot |\Vec{V}|$, where $\Vec{V}$ is the embedding vector.

\begin{table}[ht]
	\centering
	\vspace{-0.1in}
	\begin{tabular}{|c||c|c|c|}
		\hline
		Orthogonality threshold            & 0.01  &0.1 & 0.5       \\ \hline \hline
		Number of candidate words examined & $\infty$ & 372&185\\ \hline
		Model selection accuracy (\%)      & - & 96.4& 60.7\\ \hline		
	\end{tabular}
	\vspace{0.05in}
	\caption{Different values of the threshold for sub-vector orthogonality}	
	\label{tab:orthog_thr}
	\vspace{-0.35in}
\end{table}

\begin{table}[ht]
	\centering
	\vspace{-0.1in}
	\begin{tabular}{|c||c|c|c|}
		\hline
		Decomp. threshold as a fraction of $|\Vec{V}|$  & 0.02  &0.2 & 0.6       \\ \hline \hline
		Number of candidate words examined & $\infty$ & 372&78\\ \hline
		Model selection accuracy (\%)      & - & 96.4&14.3\\ \hline		
	\end{tabular}
	\caption{Different values of the threshold for decomposing the embedding vector into sub-vectors}	
	\label{tab:epsilon}
	\vspace{-0.25in}
\end{table}

\vspace{-0.05in}
\subsection{Validating the Explainable Latent Space}

As stated in Section 3, XPerT builds on the assumption that closer distances in the explainable latent space correspond to more similar language styles. We verify this assumption with a synthetic dataset that ensembles language styles at different levels. For instance, for a language style of ``expertise'', the dataset includes samples of four levels: elementary school students, middle school students, undergraduates, and PhD students. We then test the assumption by checking if the coordinates of two consecutive levels are closer.

\begin{figure}[ht]
	\centering
	\vspace{-0.05in}
	\includegraphics[width=0.95\columnwidth]{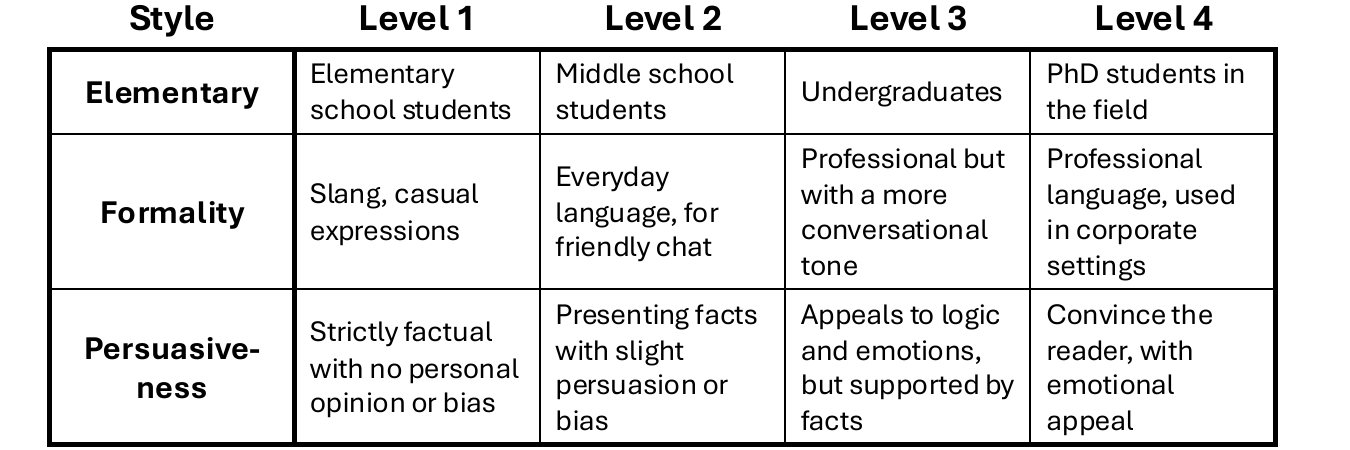} 	
	\vspace{-0.1in}
	\caption{Language styles with multiple levels}
		\vspace{-0.1in}
	\label{fig:multi_level_style}
\end{figure}

\begin{figure}
	\centering
	\subfigure[1B summarization LLM] { 
		\includegraphics[width=0.37\columnwidth]{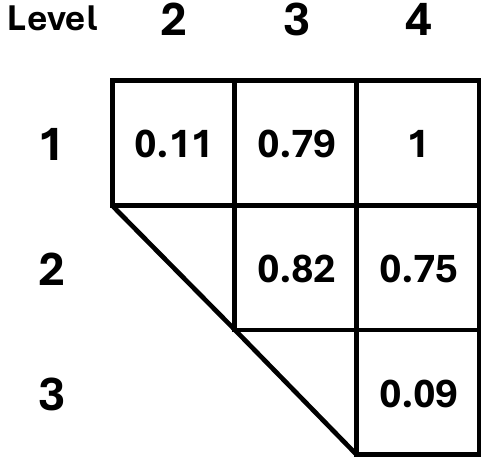}
		\label{fig:multi_level_distance_model1}
	}
		\hspace{0.2in}
	\subfigure[8B summarization LLM] { 
		\includegraphics[width=0.37\columnwidth]{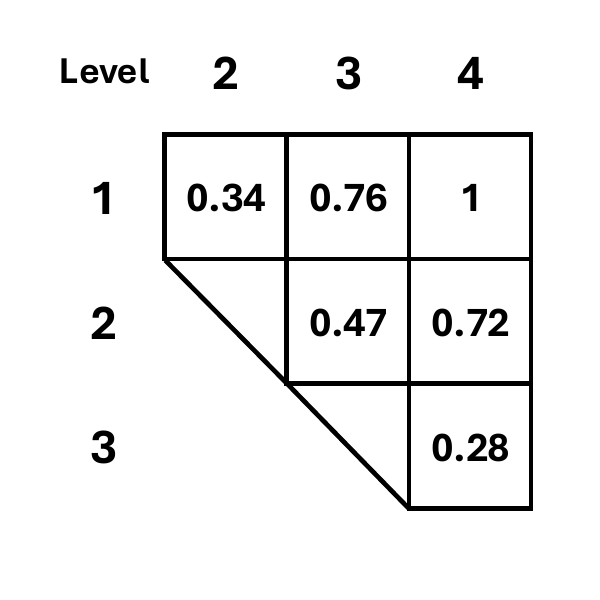}
		\label{fig:multi_level_distance_model2}
	}
	\hspace{-0.3in}
	\vspace{-0.2in}
	\caption{Distance between pLLMs coordinates fine-tuned with data at different levels of a language style}
	\label{fig:multi_level_distance}
	\vspace{-0.15in}
\end{figure}

We prompt ChatGPT4-o1 to synthesize this dataset with language styles shown in Figure \ref{fig:multi_level_style}, and then use data samples of different levels of a language style to train different pLLMs. The coordinates of these pLLMs in the latent space are then extracted from two summarization LLMs (Llama3.2-1B and Llama3.1-8B), and their distances are measured as the L1-norm. Results in Figure \ref{fig:multi_level_distance} show that the consistency between language styles and coordinates in the latent space relates to the capability of summarization LLM. With Llama3.2-1B as the summarization LLM, we can observe noticeable inconsistency, which however, can be effectively eliminated by using the Llama3.1-8B as the summarization LLM. 

	\vspace{-0.05in}
\subsection{On-Device Model Merging}

\begin{figure}[h]
	\centering
	\vspace{-0.1in}
	\includegraphics[width=0.75\columnwidth]{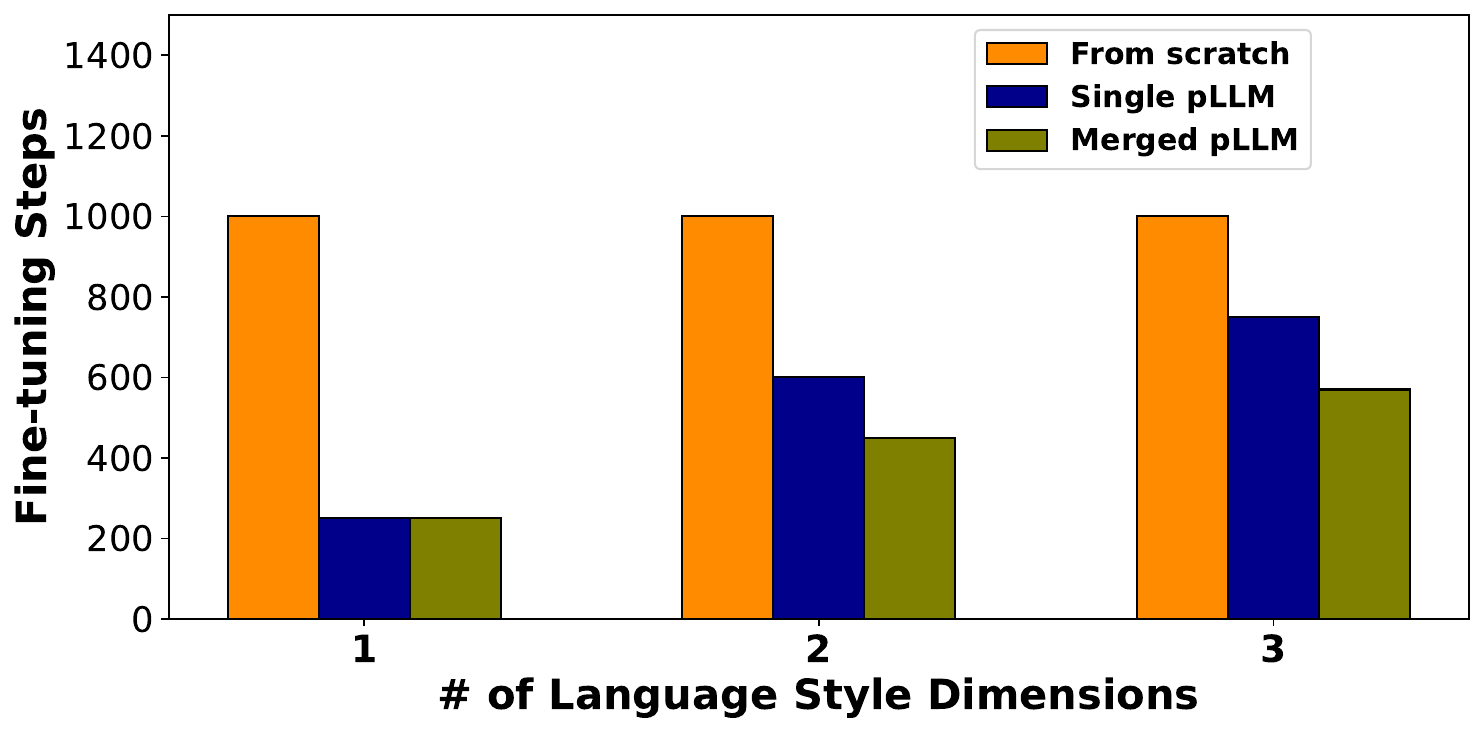} 	
	\vspace{-0.15in}
	\caption{Effectiveness of on-device model merging}
	\vspace{-0.15in}
	\label{fig:merging}
\end{figure}

It is possible that none of the cached pLLMs can well match the user's personal data, and Section 5.2 suggested to create a new pLLM with better matching by merging multiple pLLMs. To investigate the effectiveness of such on-device model merging, we consider the user's personal data as combinations of different language styles, so that none of the cached pLLM that represents one language style can well match the personal data. Results in Figure \ref{fig:merging} show that, when using Llama3.2-1B as the pLLM, as long as the user's personal data contains more than one language style, using the merged pLLM always gains extra reduction on the cost of on-device LLM personalization, by up to 25\%. 


\subsection{XPerT on Other Data Modalities}
\begin{figure}
	\centering
	\includegraphics[width=0.9\columnwidth]{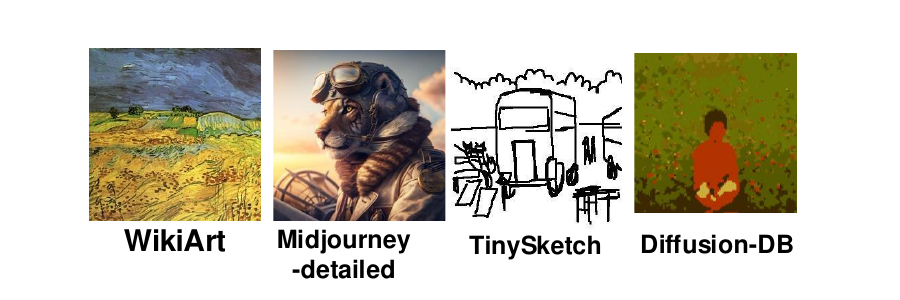} 	
	\vspace{-0.15in}
	\caption{Examples of images in different datasets with diverse artistic styles}
	\vspace{-0.15in}
	\label{fig:image_examples}
\end{figure}

To demonstrate the generalizability of XPerT, we further apply XPerT to the modality of image data. More specifically, we target the image generation task using text-to-image diffusion models, and fine-tune the U-Net structures of the Stable Diffusion v2.1 model \cite{Rombach_2022_CVPR} using the Adam optimizer with a learning rate of 1e-4. In such fine-tuning, we use multiple real-world image datasets, including WikiArt \cite{wikiart}, DiffusionDB \cite{wangDiffusionDBLargescalePrompt2022}, Midjourney-Detailed \cite{midjourney} and TinySketch \cite{fscoco}\footnote{For datasets without textual descriptions of images such as WikiArt \cite{wikiart}, we use a pre-trained BLIP2 image captioning model \cite{li2023blip} to generate the textual descriptions for image samples.}, to incorporate 9 distinct artistic image styles as exemplified in Figure \ref{fig:image_examples}, and randomly select 500 images in each style for fine-tuning the diffusion model.

In such experiments on fine-tuning the diffusion model, for XPerT, the summarization LLM must not only be capable of converting differences in images into text, but also generating synthetic image data to compute sub-vectors as described in Section \ref{subsec:decomposition_vector}. However, most open-source models that meet such multimodal requirements have only limited representation power. As a result, we use two separate models: the Qwen2-VL-7B-Instruct model \cite{Qwen2VL} for summarizing the differences in images, and the Stable Diffusion v3.5 model \cite{esser2024scaling} for generating synthetic image data.
	
\begin{figure}[h]
	\centering
	\vspace{-0.1in}
	\includegraphics[width=0.75\columnwidth]{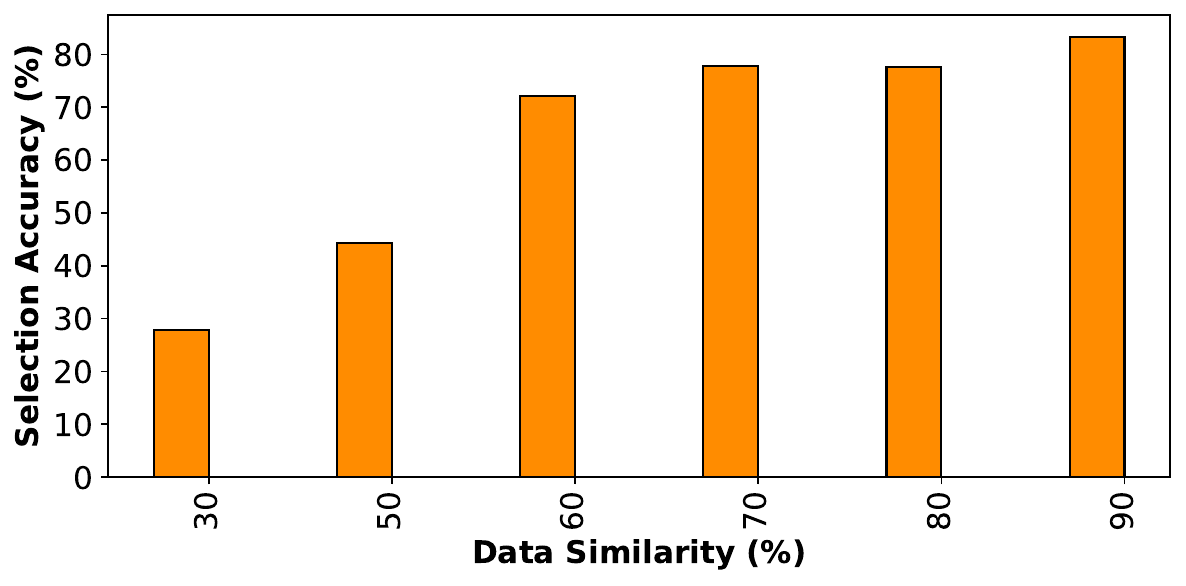} 	
	\vspace{-0.1in}
	\caption{Accuracy of pLLM selection on the image generation task}
	\vspace{-0.25in}
	\label{fig:image}
\end{figure}

Experiment results are shown in Figure \ref{fig:image}. XPerT achieves over 77.8\% accuracy of pLLM selection, when the similarity between the device's local data and the training data used in generating pLLMs exceeds 70\%. Besides, we also implemented the baseline methods by using the FID score \cite{heusel2017gans} to measure the distance between the pLLM's training data and the user's local data. Similar to results in the text modality, the selection cost of baseline methods is proportional to the number of pLLMs and is hence much higher than that of XPerT, which is independent from the number of pLLMs.


\subsection{Comparing XPerT with In-Context Learning} 

Besides fine-tuning, another option to improve the model's performance for downstream tasks is in-context learning (ICL), which allows the model to learn tasks from the few in-context examples given through the prompt \cite{brown2020language,wu2022self,xue2024phyt2v,song2024achieving}. However, compared to fine-tuning, ICL usually exhibits limited capability in complex or domain-specific tasks, due to the absence of model training efforts that are essential to teach new concepts and knowledge to the model \cite{treutlein2024connecting}. For example, ICL has been found to perform significantly worse than fine-tuning on instruction-tuning datasets, especially on small LLMs with less than 1B parameters \cite{ponce2025context}. 

To further verify this, we compared XPerT with ICL by using the Qwen2.5-1.5B-Instruct model \cite{qwen2.5} and 4 domain-specific datasets. These datasets include: 1) \emph{MedDialog} \cite{chen2020meddiag}, which contains 0.26 million dialogues between doctors and patients from healthcaremagic.com and icliniq.com; 2) \emph{DISC-Law-SFT} \cite{yue2023disclawllm}, a QA dataset covers different legal scenarios; 3) \emph{Educhat-sft} \cite{educhat-sft}, which contains 4M QA samples in education domain; and 4) \emph{FinanceAlpaca} \cite{financealpaca}, which contains 70k QA samples collected from multiple datasets in the finance domain. For XPerT, we use each dataset to fine-tune the model using the same settings as above. For ICL, we assume that users are not experts and lack the professional knowledge to select and curate ICL examples, as described in \cite{lin2023unlocking}. Instead, they merely adopt vanilla ICL with randomly chosen examples. We set the number of in-context examples to 3, and run the experiment 5 times for each dataset to obtain the averaged results. 

Experiment results in Table \ref{tab:icl} show that the performance of ICL is significantly lower than that of XPerT in different data domains. Specifically, in tasks requiring domain-specific knowledge such as the medical domain, ICL learns very little from in-context samples, resulting in substantially lower performance. In contrast, for the education domain which does not require extensive domain-specific knowledge, ICL can achieve a non-trivial improvement by learning the format from in-context examples. However, its performance is still lower than that of XPerT.

\begin{table}[ht]
	\centering
	\begin{tabular}{|c||c|c||c|c||}
		
		\hline
		
		Metric&\multicolumn{2}{c||}{\textbf{BLEU}} & \multicolumn{2}{c||}{\textbf{ROUGE-1}} \\
		\hline
		\hline
		&ICL&XPerT&ICL&XPerT\\
		\hline
		{MedDialog} \cite{chen2020meddiag} &0.042&0.128&0.069&0.152\\
		\hline
		{DISC-Law-SFT} \cite{yue2023disclawllm}&0.156&0.239&0.078&0.308\\
		\hline
		{Educhat-sft} \cite{educhat-sft}&0.229&0.255&0.207&0.332\\
		\hline
		{FinanceAlpaca} \cite{financealpaca}&0.137&0.278&0.114&0.289\\
		\hline
		
	\end{tabular}
	\vspace{0.05in}
	\caption{Performance of XPerT and ICL in different domains}  
	\label{tab:icl}
	\vspace{-0.3in}
\end{table}

Additionally, ICL's performance in instruction-following tasks is highly dependent on the specific samples used in the prompt \cite{zhao2024context}, making it unreliable in practice, particularly for non-expert users. Beyond its limited capability, ICL requires the model to process additional tokens and incurs extra computational cost in inference. 

\vspace{-0.05in}
\section{Related Work}

\noindent\textbf{Linear representation in LLM.} Well-trained LLMs are believed to encode linear representations of human-interpretable concepts, such as gender \cite{mikolov2013linguistic,elhage2022toy, wang2024concept,nanda2023emergent}. If the direction of a concept in the embedding space is known, the semantics of the concept representation can be adjusted using simple linear operations. 
This linearity was also observed in other domains including computer vision models \cite{wang2023concept,trager2023linear} and intelligent systems \cite{schut2023bridging}.  Hence, in XPerT we build on this observation and achieve explainability of the embedding vector by linearly decomposing it using concept representations.

\noindent\textbf{Extraction of text features.} Our main approach in XPerT is the leverage the natural language capability of LLM to encode the difference between pLLMs' outputs into the embedding vector. On the other hand, to convert LLMs' output texts into such a feature vector, the simplest way is to use a pre-trained text encoder, such as CLIP \cite{radford2021learning}. However, most existing text encoders are trained to extract the text's semantic meaning \cite{kim2024fine} rather than high-level concepts such as the language style. An alternative approach is to train a language style classification model that can directly extract language style features, but the existing pre-trained models are limited to a small number of language styles \cite{jin2022deep}, such as formality and toxicity. Additionally, the feature vectors extracted by these methods lack sufficient explainability.

\noindent\textbf{Model selection.} Our design of XPerT is related to the existing work in model selection, which selects the model from a set of candidates that best satisfies the task requirement or user context \cite{foster2017parameter,foster2019model,li2020efficient,karimi2021online}. The candidates could be models with different architectures or the same model with different hyperparameters. 
In our scenario, the candidate models are the same LLM with different initial weights, which can also be seen as a kind of hyperparameter. Our design of XPerT, however, is fundamentally different because we extract high-level information from each candidate model for selection, instead of directly evaluating the candidate's performance.

\vspace{-0.1in}
\section{Discussions}


\noindent\textbf{Other ways of personalization.} Besides different language styles, an LLM can also be personalized in many other aspects. For example, a code generation LLM can be personalized for different programming languages \cite{cambaz2024use}, and a QA LLM can be customized for specific domains such as history or physics \cite{anand2024mm}. XPerT can also be applied to these aspects, by tailoring the prompt in Figure \ref{fig:prompt_template} accordingly. Unlike stylized text generation, tasks of code generation and domain-specific QA require more explicit methods of reflecting the personalized aspects in the LLM outputs, and it is better the prompt the summarization LLM with more task-specific instructions.

\noindent\textbf{Untrusted cloud sources.} Users may hesitate to use a pLLM downloaded from the cloud for on-device personalization, because such as download pLLM from untrusted cloud sources may be backdoored by malicious model publishers \cite{li2024backdoorllm}. To mitigate such risks, the users should be advised to only use pLLMs from trusted cloud sources. Furthermore, fine-tuning a backdoored LLM with benign data (i.e., the user's personal data) has been verified to effectively mitigate the backdoor attack \cite{liu2017neural}, and lightweight backdoor detection methods are also available for use on mobile devices \cite{qi2020onion}.


\noindent\textbf{Generalizability of personalized LLM}. We evaluate the model's performance only on datasets of specific downstream tasks, rather than on standard benchmarks for general assessment. This is because, by definition, fine-tuning improves the model's performance on a specific downstream task. Hence, generalizability is naturally reduced in a fine-tuned model compared to the pre-trained model, and this loss of generalizability is orthogonal to XPerT's design.

\noindent\textbf{Connectivity to the server}. In XPerT, the user's device needs to download the selected pLLM from the server, and we assume that mobile devices have an almost always-on connection to the server via wireless links. This assumption is the cornerstone of many smartphone services and apps, including cloud storage, AI assistants, and streaming platforms. Additionally, XPerT only communicates with the server once to download the pLLM; afterwards, local fine-tuning can proceed without any further communications.

\vspace{-0.1in}
\section{Conclusion}
In this paper, we present XPerT, a new technique that expedites on-device LLM personalization by fine-tuning the proper pLLM cached at the cloud server with on-device personal data. XPerT provide detailed and quantitative explainability of pLLMs, so as to ensure proper model selection.

\vspace{-0.1in}
\section*{Acknowledgments}
We thank the shepherd and reviewers for their comments and feedback. This work was supported in part by National Science Foundation (NSF) under grant number IIS-2205360, CCF-2217003, CCF-2215042, and National Institutes of Health (NIH) under grant number R01HL170368.

\bibliographystyle{ACM-Reference-Format}
\bibliography{mobisys25}

\end{document}